\documentclass{article}

\usepackage[final]{neurips_2025}

\usepackage[utf8]{inputenc} 
\usepackage[T1]{fontenc}    
\usepackage{hyperref}       
\usepackage{url}            
\usepackage{booktabs}       
\usepackage{amsfonts}       
\usepackage{nicefrac}       
\usepackage{microtype}      
\usepackage[table]{xcolor}
\usepackage{colortbl}
\usepackage{amsmath}
\usepackage{graphicx} 
\title{Dual-Stage Value-Guided Inference with Margin-Based Reward Adjustment for Fast and Faithful VLM Captioning}
\usepackage{algorithm}
\usepackage[noend]{algpseudocode}

\usepackage{multirow}
\usepackage{caption}
\usepackage{geometry}
\usepackage{tcolorbox}
\usepackage{siunitx}
\definecolor{highlight}{HTML}{FEEEE1}  
\usepackage{subcaption}          

\usepackage{tikz} 
\newcommand{\circled}[1]{\tikz[baseline=(char.base)]{
            \node[shape=circle,draw,inner sep=1pt] (char) {\textbf{#1}};}}

\author{
\textbf{Ankan Deria}$^{1}$, \textbf{Adinath Madhavrao Dukre}$^1$, \textbf{Feilong Tang}$^1$, \textbf{Sara Atito}$^{2}$, \textbf{Sudipta Roy}$^3$ \\ \textbf{Muhammad Awais}$^{2}$, \textbf{Muhammad Haris Khan}$^{1}$, \textbf{Imran Razzak}$^{1,4}$ \And 
\textnormal{$^1$ Mohamed bin Zayed University of AI, Abu Dhabi, UAE } \\
\textnormal{$^2$ University of Surrey, UK} \\
\textnormal{$^3$ Jio Institute, India} \\
\textnormal{$^4$ UNSW, Australia}
}

\begin{document}

\maketitle

\begin{abstract}

Despite significant advances in inference-time search for vision–language models (VLMs), existing approaches remain both computationally expensive and prone to unpenalized, low-confidence generations which often lead to persistent hallucinations. We introduce \textbf{Value-guided Inference with Margin-based Reward (ViMaR)}, a two-stage inference framework that improves both efficiency and output fidelity by combining a temporal-difference value model with a margin-aware reward adjustment. In the first stage, we perform a single pass to identify the highest-value caption among diverse candidates. In the second stage, we selectively refine only those segments that were overlooked or exhibit weak visual grounding, thereby eliminating frequently rewarded evaluations. A calibrated margin-based penalty discourages low-confidence continuations while preserving descriptive richness. Extensive experiments across multiple VLM architectures demonstrate that ViMaR generates captions that are significantly more reliable, factually accurate, detailed, and explanatory, while achieving over 4$\times$ speedup compared to existing value-guided methods. Specifically, we show that ViMaR trained solely on LLaVA Mistral-7B, \textit{generalizes effectively to guide decoding in a stronger unseen model}. To further validate this, we adapt the ViMaR to steer generation in LLaVA-OneVision-Qwen2-7B, leading to consistent improvements in caption quality and demonstrating robust cross-model guidance. This cross-model generalization highlights ViMaR's flexibility and modularity, positioning it as a scalable and transferable inference-time decoding strategy. Furthermore, when ViMaR-generated captions are used for self-training, the underlying models achieve substantial gains across a broad suite of visual comprehension benchmarks, underscoring the potential of fast, accurate, and self-improving VLM pipelines.

\end{abstract}

\vspace{-5pt}
\section{Introduction}
\label{introduction}
\vspace{-5pt}
Vision–language models (VLMs) \cite{bai2023qwen,beyer2024paligemma,chen2024internvl,li2023blip,liu2024llavanext} have revolutionized our ability to produce fluent, richly detailed image descriptions.  However, they still contend with two intertwined challenges: generating precise, fine-grained captions and combating “hallucinations”~\cite{liu2023mitigating, beyer2024paligemma, wang2024mementos, wang2024scaling, liu2024llavanext}, which often arise from unpenalized, low-confidence outputs that misrepresent the scene. Simply scaling up training data can ameliorate these issues, but it incurs prohibitive annotation and API costs, making it fundamentally unscalable.  Meanwhile, standard decoding strategies such as best-of-\(N\) or greedy sampling either underutilize the model’s representational power or demand exhaustive, compute-intensive re-scoring of every candidate at each generation step \cite{song2024good, wan2023faithfulness}.

Recent advances in large language models (LLMs) \cite{openai2024reasoning, snell2024scaling,yang2024qwen2} have demonstrated that inference-time search—using a pretrained process-reward model to iteratively refine candidate outputs can substantially elevate response quality and even generate synthetic data for further model training \cite{tian2024toward,zhang2024rest}. Extending this paradigm to VLMs, however, introduces unique challenges: unlike text-only tasks, VLMs require a reward signal that captures both visual grounding and linguistic coherence across multiple sentences. To address this, Wang~\emph{et al.}~\cite{wang2024scaling} proposed the Vision Value Model (VisVM), which employs temporal-difference learning~\cite{sutton1988learning} over CLIP similarity scores to estimate the long-term quality of each candidate sentence. Though, VisVM markedly reduced hallucination and enriches visual details, however scoring every candidate at each step incurs considerable overhead. 



To overcome aforementioned challenges, we present a two-stage search pipeline that preserves or improves caption fidelity while reducing end-to-end inference time by over $4×\times$. In the first stage, a single 'best-of' pass selects the highest-value caption (coarse caption) under our trained policy. In the second stage, we propose to perform targeted refinement (fine caption): additional search is restricted solely to those segments where salient image regions were likely overlooked, thereby eliminating the need to re-score the full candidate set on every generation step.  At the core of our method lies a new margin-based reward adjustment for training: whenever a candidate’s CLIP similarity falls below a calibrated threshold, we impose a penalty proportional to the gap. This mechanism seamlessly integrates into the existing temporal difference framework, sharpening the model’s preference for factually grounded, detail-rich phrases. In results, ViMaR demonstrates strong cross-model generalization: despite being based on LLaVA Mistral-7B, ViMaR effectively guides decoding on other stronger models such as LLaVA OneVision-Qwen2-7 B. This highlights the generalizability and scalability of our framework, making it suitable as a plug-and-play decoding strategy across diverse VLM architectures.  Furthermore, by leveraging ground-truth captions in our dataset, we ensure that truly accurate descriptions receive the highest reward signals. We validate our approach in two comprehensive studies. First, in COCO-based descriptive captioning, our optimized value-guided search produces captions that are markedly more detailed and substantially less prone to hallucination than those from VisVM search, greedy decoding, best-of-\(N\) or standard CLIP-PRM search. In blind evaluations conducted using both GPT-4o and human judges, our captions are preferred in 49.3\% and 64\% of pairwise comparisons, respectively, against outputs generated by the VisVM (see Figure \ref{fig:two_gpt_wins}). These performance gains come at a fraction of the cost: our inference pipeline runs significantly faster than state-of-the-art VisVM. Second, we harness these high-quality captions to self-train the base LLaVA-Next-7B model.  Fine-tuning on our optimized-generated data yields consistent improvements across eight diverse multimodal benchmarks, achieving an average performance uplift of 15.87\%.  Together, these results highlight the potential of a fast, accurate, and computationally efficient VLM inference paradigm that supports self-improvement. 



\section{Related Work}
\label{sec:related_work}

\noindent\textbf{Vision–Language Modeling:}Early joint vision–text models combined convolutional or transformer-based image encoders with sequence decoders to tackle tasks such as object tagging, image captioning, and visual question answering~\cite{guo2023images, li2019coco, yu2022coca}.  More recent approaches fuse large pretrained language backbones with powerful visual representations (e.g., CLIP) to enable instruction following, in-context multimodal reasoning, and zero-shot generalization~\cite{alayrac2022flamingo, chen2024internvl, wang2024qwen2}.  Despite these capabilities, VLMs remain prone to \textit{hallucination}, producing confidently stated but incorrect content~\cite{bai2024hallucination, guan2024hallusionbench, rohrbach2018object}.

\noindent\textbf{Hallucination Mitigation:}
Hallucinations in vision–language models are typically addressed in the following ways. Enhance the quality of supervised fine-tuning (SFT) datasets through human annotation, synthetic caption rewrites, or contrastive filtering to provide more accurate grounding~\cite{chen2023mitigating, chen2024sharegpt4v, wang2024cogvlm, wang2024mitigating}, or apply corrective methods during post-training, such as fine-tuning with adversarial negatives, consistency checks, or calibrate self-rewarding to detect and suppress spurious phrases~\cite{liu2023mitigating, sun2023aligning, zhou2024calibrated, luo2024hallucination}.  In contrast, our approach leaves model weights and training data unchanged, instead devising a two-stage inference-time search to actively reward well-grounded descriptions while penalizing low-confidence outputs that are prone to hallucination.


\noindent\textbf{Descriptive Paragraph Captioning:}
Descriptive captioning extends single-sentence models to produce multi-sentence paragraphs that comprehensively describe both global scene context and fine-grained regional details~\cite{chen2024internvl, betker2023improving}.  Early work on paragraph captioning demonstrated that standard sequence models often generate repetitive text with limited diversity, motivating approaches that explicitly promote novel content and discourage redundancy~\cite{krause2017hierarchical, melas2018training}.  Reinforcement learning methods such as Self-Critical Sequence Training have been adapted to optimize non-differentiable paragraph-level metrics, improving coherence but still requiring heavy sampling during inference~\cite{rennie2017self}.  Partially non-autoregressive architectures further reduce latency by updating only segments of the caption in parallel, yet they can struggle to maintain sequential consistency across sentences \cite{fei2021partially}. To mitigate low-quality outputs, DeepSeek VL2 employs a lightweight quality control pipeline powered by DeepSeek Chat to quickly score and filter captions based on writing quality alone, effectively pruning imprecise or bland descriptions before post-training ~\cite{wu2024deepseek, liu2024deepseek}.  More recent strategies incorporate human-style feedback, for example, caption reformulations, to refine output at inference time, achieving gains in factuality and readability without additional supervision~\cite{berger2025improving}.  Nevertheless, these methods typically rerank or regenerate full paragraphs at each step, incurring substantial compute.  In contrast, our approach targets the inference-time search itself: By unifying a two-stage best-of pass with targeted segment refinement and a margin-based penalty for low-confidence phrases, we obtain richly detailed paragraphs with minimal extra computation.


\noindent\textbf{Inference-Time Search:}
Inference-time search has emerged as a powerful mechanism for enhancing model outputs in domains ranging from code generation and mathematical reasoning to multimodal planning and robotics~\cite{silver2016mastering, uesato2022solving, zhang2024rest, chakraborty2024transfer}.  In the text-only setting, techniques such as controlled decoding~\cite{chakraborty2024transfer}, Best-of-\(N\) sampling~\cite{brown2024large, lightman2023let}, and Monte Carlo Tree Search~\cite{tian2024toward, wang2024litesearch} have demonstrated consistent gains by leveraging a learned process or value network to rerank multiple candidate continuations. During inference-time search, an effective process reward model (PRM) is crucial, since the fidelity of its reward signals directly impacts both the quality of the generated responses and the computational resources required to achieve them.

Translating these ideas to vision–language models (VLMs) introduces unique challenges: the reward signal must capture both visual–text alignment and sequential coherence across sentences.  Zhou \emph{et al.}~\cite{zhou2024calibrated} first explored CLIP-based scoring as a proxy reward, using positive and negative sample mining to refine the model post hoc.  Xiong \emph{et al.}~\cite{xiong2024llava} proposed LLAVA-Critic, which evaluates entire paragraph-level captions to filter out poor outputs, but lacks stepwise granularity.  Most recently, Zhang \emph{et al.}~\cite{zhang2024rest} introduced Rest-MCTS*, employing process-reward–guided tree search to iteratively refine multimodal responses, yet still incurs a quadratic inference cost as the tree grows.
Building on temporal-difference value learning in VLMs, Wang \emph{et al.}~\cite{wang2024scaling} presented the Vision Value Model (VisVM), which estimates both immediate and future sentence value via CLIP similarity and steers search toward low-hallucination, high-detail trajectories.  Although VisVM substantially elevates descriptive quality and reduces hallucinations, its naïve implementation must re-score all \(N\) candidates at each generation step, resulting in an \(O(N \times S)\) inference overhead (where \(S\) is the number of sentences in each step).  
Inspired by these advances, we developed \textbf{ViMaR}, a two-stage inference framework that preserves long-term value signals while improving efficiency. ViMaR conducts a best-of pass followed by targeted refinement, applies a margin-based penalty to reduce redundant scoring of every candidates, and uses beam search for stable, diverse decoding~\cite{brown2024large}. This achieves over \(4\times\) faster inference while maintaining or improving caption quality.


\vspace{-7pt}
\section{Value Guided Inference Framework-ViMaR} 
\vspace{-5pt}
We formulate the VLM captioning process as a sequential generation task modeled by a policy \(\pi_\theta\) over a probability distribution \(p_\theta\). Given an input pair consisting of a textual prompt \( x \) and an image \( I \), the model produces a multi-sentence caption \( y = [y_0; y_1, y_2, \ldots, y_m] \), where $y_0$ is the first step caption and each \( y_{i>0} \) denotes a sentence-level output. At the first step, the model produces $y_0$ by sampling from  $y_0 \sim p_\theta\bigl(\,\#\mid x, I\bigr),$ while each subsequent sentence $y_{i>0}$ is drawn conditionally from $y_{i>0} \sim p_\theta\bigl(\,\cdot\mid x, I, y_{<i}\bigr)$, followed by evaluation and potential selection at each step.
We cast this caption generation process as a Markov Decision Process (MDP) defined by the tuple \((\mathcal{S}, \mathcal{A}, \mathcal{R}, \gamma)\), where each state \(s_i \in \mathcal{S}\) consists of the prompt–image pair \((x, I)\) and the sequence of previously generated sentences \(y_{<i}\), and the action \(y_i \in \mathcal{A}\) transitions the model to the next state \(s_{i+1}\). The reward function \(\mathcal{R}(s_i, y_i)\), parameterized by a value model \(V_\rho\), scores the quality of the generated output at each step, while the discount factor \(\gamma \in [0, 1]\) governs the trade-off between immediate and future rewards. This MDP formulation enables inference-time search to explore alternative trajectories and prioritize high-quality, visually grounded captions through value-guided decoding.

\subsection{ViMaR Training}
\label{sec:td_training}
\paragraph{Training Method: }  Our proposed model, ViMaR, is designed to estimate the long-term utility of image-conditioned sentence candidates, accounting for their potential to influence subsequent generation steps. We adopt a temporal-difference (TD) learning strategy~\cite{sutton1988learning}, which enables ViMaR to recursively refine its predictions of the cumulative reward from any given state $s_i = (y_i, I)$, where $y_i$ is the current sentence and $I$ is the input image.

Given a training triplet $(y_i, y_{i+1}, I)$—comprising the current and next sentence in a paragraph, along with the associated image—we begin by computing the similarity score $\delta$ between the sentence $y_i$ and the image $I$ using a pretrained process reward model (PRM). To incorporate a mechanism that discourages low-confidence, potentially hallucinatory outputs, we introduce a margin-based reward adjustment. The reward $r_{s_i}$ at each state is computed as:
\begin{equation}
    r_{s_i} =
    \begin{cases}
        \delta, & \text{if } \delta \ge \tau, \\
        \tau - \delta, & \text{otherwise}
    \end{cases}
\end{equation}
Here, $\tau$ denotes a calibrated threshold that serves as a margin for penalizing uncertain or weakly grounded predictions. When the PRM score falls below this threshold, a negative penalty proportional to the margin gap is applied, encouraging the model to avoid such candidates during search.

The model is trained to minimize the discrepancy between the predicted value of the current state and the target value, which is defined as the sum of the immediate reward and the discounted value of the next state. Formally, the training objective is:
\begin{equation}
    \mathcal{L}(\rho) = \mathbb{E}_{(y_i, y_{i+1}, I) \sim \mathcal{D}} \left[ \left( r_{s_i} + \gamma V_\rho(y_{i+1}, I) - V_\rho(y_i, I) \right)^2 \right]
\end{equation}
Here, $V_\rho$ is the value predicted by ViMaR, $\gamma$ is the discount factor, and $\rho$ denotes the learnable model parameters. The training set $\mathcal{D}$ comprises image-caption pairs segmented into sentence-level transitions to capture both local grounding and long-term contextual dependencies.
\paragraph{Training Data:}
\label{sec:vm_training}
To train ViMaR, we construct training triplets of the form \((y_i, y_{i+1}, I)\), where \(y_i\) is a sentence from a paragraph-level caption, \(y_{i+1}\) is its immediate successor, and \(I\) is the corresponding image. These triplets are derived from multi-sentence image descriptions \(y = [y_1, y_2, \ldots, y_m]\) paired with their respective images. Modeling the long-term value of a sentence requires capturing not only its direct alignment with the image but also its downstream influence on the continuation of the caption. To ensure a diverse set of generation patterns, we begin with 23K images from the COCO 2017 training split and pair them with detailed prompts from the LLaVA-150K dataset. For each image–prompt pair, we include both the ground-truth caption and five additional captions generated by a VLM using a mix of greedy decoding and temperature-controlled sampling to promote diversity. Each paragraph is then segmented into ordered sentence pairs, yielding a total of 792K triplets. We used 732K examples for training and 60K for validation. 
\paragraph{Implementation Details:}  
We build ViMaR on top of the LLaVA-Next-Mistral-7B architecture. Concretely, we attach a linear value head to the penultimate transformer layer; this head outputs a scalar estimate of the cumulative, long-term reward for each image–sentence state. All other weights in ViMaR are initialized from the pretrained LLaVA-Next-Mistral-7B checkpoint and remain trainable alongside the new value head.
For the process-reward model (PRM), we choose the CLIP-ViT. This choice offers two advantages: (1) CLIP’s image–text embedding similarity provides a proven metric for visual grounding, yielding reliable reward signals for descriptive captioning; and (2) leveraging the native CLIP-ViT avoids external dependencies or costly human annotations, creating a fully self-contained training pipeline and easily customized with our margin-based reward adjustment. To support penalization and reduce hallucination, we modify the CLIP-based PRM by introducing a margin-based reward adjustment (as described in Section~\ref{sec:td_training}), thereby downweighting low-confidence alignments during reward computation.




\subsection{Inference-Time Search with Two-Stage Refinement}
\label{sec:two_stage_search}

Once trained, ViMaR serves as a value model \(V_\rho\) to guide inference-time search, enabling the VLM to produce more accurate and visually grounded descriptions. 
In the \textbf{first stage}, we perform full-paragraph generation using beam search over the entire prompt–image pair \((x, I)\), applying temperature sampling with \(N\) distinct decoding temperatures \(\{T_n\}_{n=1}^N\). For each temperature \(T_n\), the model samples \(K\) complete paragraph-level candidates from the policy:
$ y \sim p_\theta(\,\# \mid x, I, T_n), $
where \(\#\) denotes the end-of-caption token. This results in a total of \(N \times K\) candidate captions. Each is scored holistically by the value model \(V_\rho(y, I)\), and the caption with the highest predicted value is selected as the base output. \textbf{In the second stage}, we perform targeted refinement on the selected base caption. For each segment \(y_i\) that lacks sufficient visual grounding—identified by low confidence or by missing objects in the image—we resample alternatives from the conditional distribution
$y_i \sim p_\theta(\cdot \mid x, I, y_{<i}, T_n),$
drawing \(K\) candidates per temperature over \(N\) temperatures for a total of \(N \times K\) alternatives. Each candidate is scored by the value model \(V_\rho(y_i, I)\), and the highest-value sentence is incorporated into the caption. This refinement loop repeats until all salient content is addressed and an end-of-sequence (EOS) token is generated. This two-stage search procedure preserves the long-horizon reasoning of the full caption while selectively improving weak segments, significantly reducing inference-time cost without sacrificing detail or faithfulness. A complete overview is provided in Algorithm~\ref{alg:search}.


\vspace{-5pt}
\begin{algorithm}[t]
\caption{Two-Stage Inference-Time Search with ViMaR}
\label{alg:search}
\textbf{Require:} Test sample $\{x, I\}$, VLM policy $p_\theta$, value model $V_\rho$, temperature list $T = \{T_n\}_{n=1}^N$, candidate count $K$\\
\textbf{Output:} Final response $y$
\begin{algorithmic}[1]
\State \textcolor{gray}{\# Stage 1: Generate diverse base captions}
\State Initialize candidate set $\mathcal{C} = \emptyset$
\For{$T_n \in T$}
    \State Generate $K$ paragraph-level responses $\{y^{(n,k)}\}_{k=1}^K \sim p_\theta(\cdot \mid x, I, T_n)$
    \State Add all $\{y^{(n,k)}\}_{k=1}^K$ to $\mathcal{C}$
\EndFor
\State Select base caption $y^{*} = \arg\max_{y \in \mathcal{C}} V_\rho(y, I)$

\State \textcolor{gray}{\# Stage 2: Add supplementary segments to improve grounding}
\State Identify under-grounded or missing visual regions in $y^{*}$
\For{while Generation is not Done}
    \State Initialize candidate set $\mathcal{S}_i = \emptyset$
    \For{$T_n \in T$}
        \State Generate $K$ candidate sentences $\{s_i^{(n,k)}\}_{k=1}^K \sim p_\theta(\cdot \mid x, I, y_{<i}, T_n)$
        \State Add all $\{s_i^{(n,k)}\}$ to $\mathcal{S}_i$
    \EndFor
    \State Select best sentence $s_i^{*} = \arg\max_{s \in \mathcal{S}_i} V_\rho(s, I)$
    \State Append $s_i^{*}$ to $y^{*}$ at the appropriate position
\EndFor
\State \Return final refined response $y^{*}$
\end{algorithmic}
\end{algorithm}

\vspace{-5pt}
\section{Experiments}
\vspace{-5pt}
In this section, we empirically evaluate ViMaR-guided inference-time search framework. Our investigation is centered around the following key questions: (1) Does the proposed two-stage decoding strategy, guided by the learned value model, generate more accurate and visually grounded outputs compared to existing inference-time decoding methods? (Section~\ref{sec:quality_eval}) (2) Can ViMaR’s refined outputs serve as high-quality supervision signals to enhance the visual comprehension capabilities of VLMs through self-training? (Section~\ref{sec:self_train}) (3) How efficient is our method in terms of inference speed compared to baseline search strategies? (Section~\ref{sec:time_efficiency})

\subsection{Evaluating the Effectiveness of ViMaR-Guided Two-Stage Search}
\label{sec:quality_eval}
\paragraph{Baselines and Implementation Details: }
We compare our two-stage ViMaR-guided search against four established inference-time decoding strategies, all built on LLaVA-Next-Mistral-7B:\textbf{Greedy Decoding:} Stepwise selection of the highest‐probability token. \textbf{Best-of-N (BoN):} Generate 30 full captions using five temperatures $\{0.1,0.3,0.5,0.7,0.9\}$ (six per temperature) and choose the best via GPT-4o. \textbf{CLIP-PRM Guided Search:} Stepwise search using CLIP–ViT similarity as the reward, with temperature decoding ($N=5$) and $K=6$ samples per temperature. \textbf{VisVM-Guided Search:} Single-stage inference-time search guided by the Vision Value Model, evaluating all $N\times K$ candidates at each sentence step. \textbf{ViMaR Two-Stage Search (Ours):} Stage 1 generates paragraph candidates with $(N=5, K=6)$ and selects the best by $V_\rho$; Stage 2 refine and add additional segments to improve the caption details.  All methods employ LLaVA-Next-Mistral-7B as the base VLM and initialize ViMaR’s value head from its penultimate layer. We fix the temperature set to $\{0.1,0.3,0.5,0.7,0.9\}$ and sample $K=6$ candidates per temperature in both stages. CLIP–ViT (the native LLaVA encoder) serves as the PRM for consistency and cost-efficiency. In all experiments, we kept the total decode calls identical to ensure a fair comparison of quality versus compute.




\noindent\textbf{\circled{1} Two-Stage Value-Guided Search Enhances Caption Quality}
\label{sec:human_gpt_eval}

\begin{figure}[ht]
  \centering
  \begin{subfigure}[b]{0.49\linewidth}
    \centering
    \includegraphics[width=0.9\linewidth]{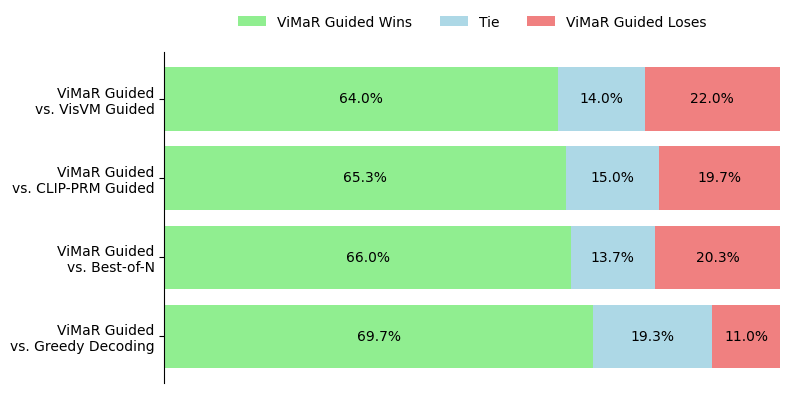}
    \caption{Human evaluation of win rates.}
    \label{fig:human_evaluation}
  \end{subfigure}
  \hfill
  \begin{subfigure}[b]{0.49\linewidth}
    \centering
    \includegraphics[width=0.90\linewidth]{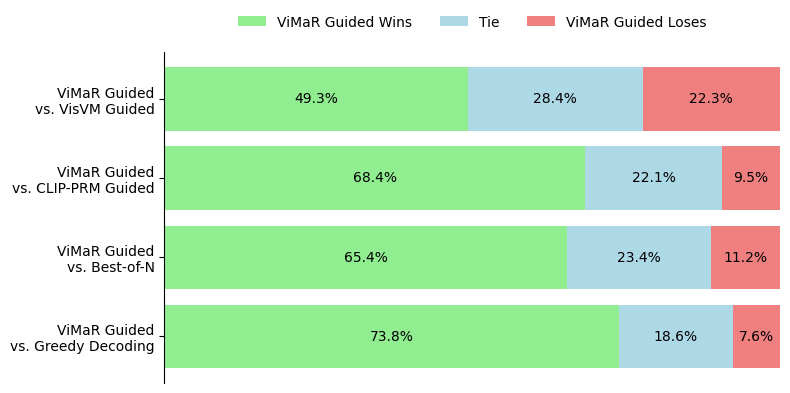}
    \caption{GPT-4o evaluation of win rates.}
    \label{fig:gpt_evaluation}
  \end{subfigure}
  \caption{Comparison of image-description quality across search strategies for LLaVA-Next-7B. (a) Independent human raters corroborate these findings, selecting ViMaR-guided outputs at significantly higher rates than all other methods. (b) Win-rate judgments by GPT-4o show that ViMaR-guided search consistently outperforms all other search methods.}
  \label{fig:two_gpt_wins}
\end{figure}

\begin{figure}
    \centering
    \includegraphics[width=0.94\linewidth]{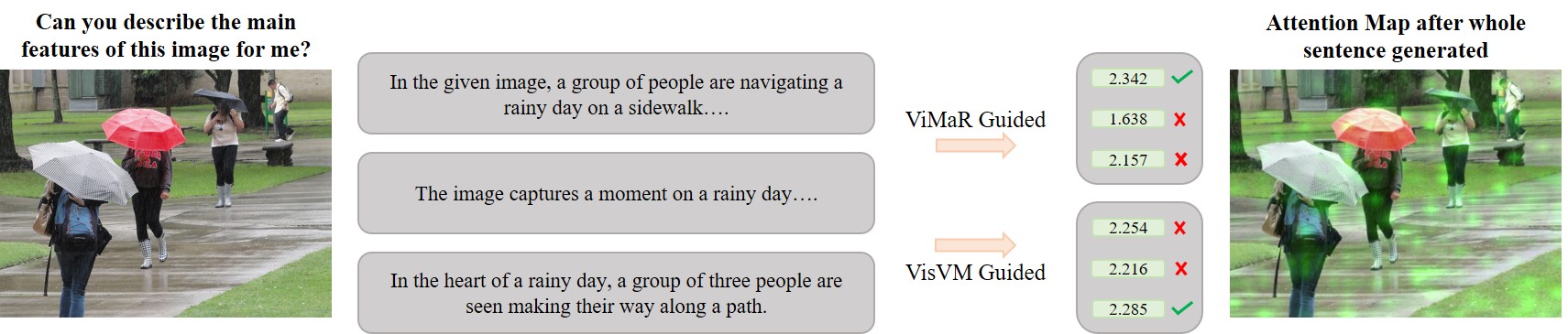}
    \includegraphics[width=0.94\linewidth]{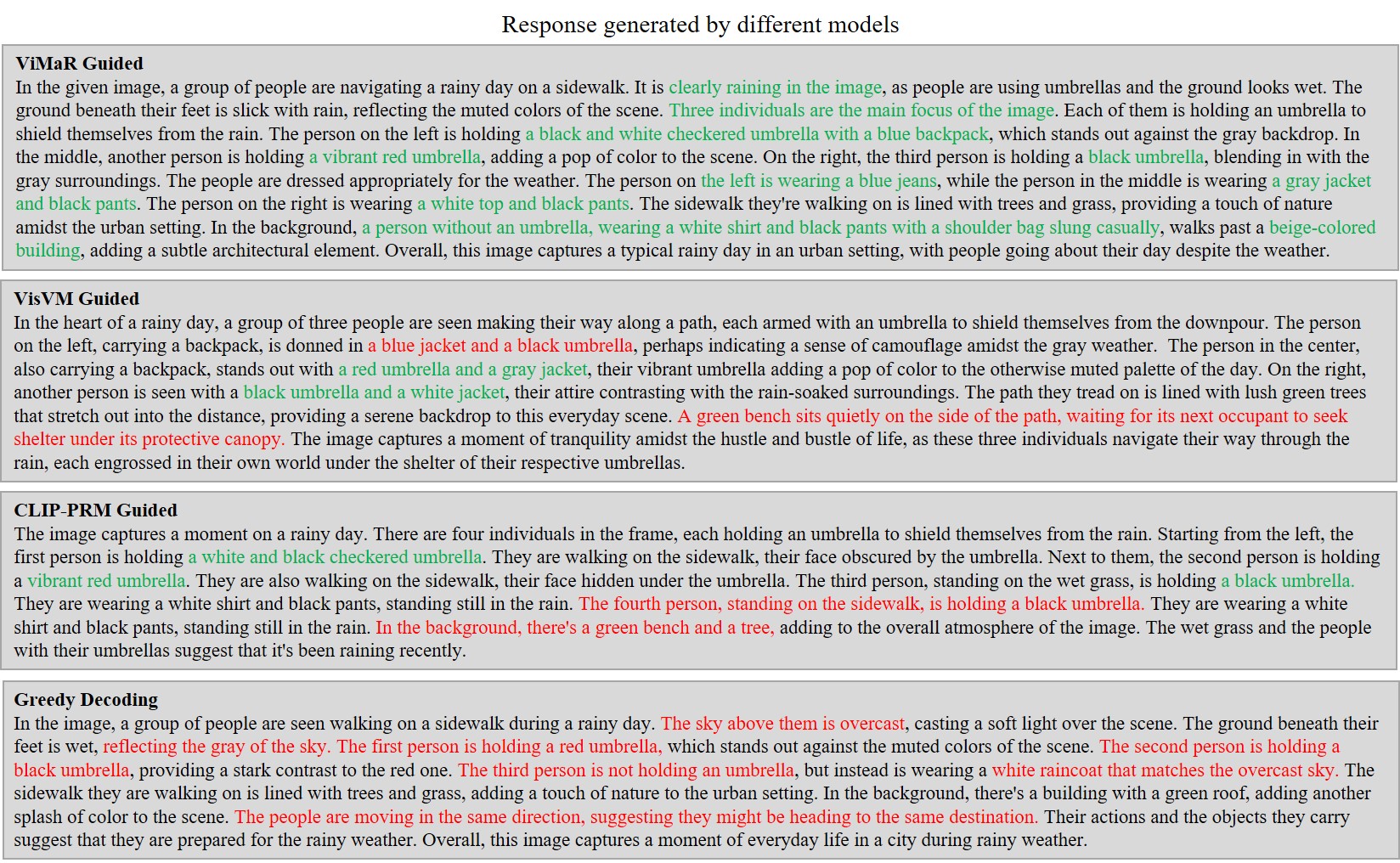}
    \caption{Qualitative comparison of decoding strategies. The top section shows how our value model and VisVM select different candidate responses, along with the resulting changes in LLaVA-Next's attention maps. Notable shifts in attention distribution highlight the influence of value-guided selection. The bottom section presents full captions generated by various search methods. Correctly grounded details are marked in \textcolor{green}{green}, while hallucinated content is highlighted in \textcolor{red}{red}. Our two-stage decoding strategy produces responses that are more accurate, detailed, and visually aligned.}
    \label{fig:observation}
\end{figure}

To evaluate the effectiveness of our proposed search strategy, we sample 1,000 images from the COCO Train2017 dataset and pair each image with the prompts from the LLaVA-150k detailed description dataset, resulting in 1,000 image–prompt pairs for evaluation. We generate one descriptive caption per pair using our two-stage value-guided decoding strategy and four alternative decoding methods—including greedy decoding, BoN search, CLP-PRM sampling, and the original VisVM-guided search. The quality of the generated captions is assessed through both human preference studies and automated metrics. For human evaluation, we randomly select 300 image–prompt pairs and ask annotators to compare outputs from our method against each baseline, identifying the preferred response in each case. As shown in Figure~\ref{fig:human_evaluation}, our two-stage strategy consistently outperforms all baselines, achieving win rates of 64.0\%, 65.3 \%, 66.0\% and 69.7\% over VisVM-guided search, CLIP-PRM, BoN and greedy search, respectively. Notably, greedy decoding performs the worst, while VisVM-guided search offers meaningful improvements—but still lags behind our approach, highlighting the benefits of long-horizon and localized refinement. As illustrated in Figure \ref{fig:observation}, our method generates descriptions that are both richer in detail and better aligned with visual content. For instance, descriptions include nuanced elements such as ``clearly raining in the image", which are often omitted by competing methods. In addition, we evaluate model outputs using GPT-4o-based pairwise comparisons. Figure~\ref{fig:gpt_evaluation} indicate that captions generated with our two-stage method are preferred in 49.3\%, 68.4\%, 65.4\%, and 73.8\% of the cases over the same four baselines. These findings demonstrate that our search strategy improves both the fidelity and richness of generated descriptions, pushing the boundaries of VLM visual comprehension.


\noindent\textbf{\circled{2} Two-Stage Value-Guided Search Mitigates Visual Hallucinations}\\
\label{sec:hallucination}

To assess the impact of our two-stage value-guided search on visual hallucination, we conduct a quantitative evaluation using 500 randomly sampled images from the COCO Val2017 dataset. Each image is paired with detailed prompts sourced from the LLaVA-150k dataset. We employ two widely used metrics to measure hallucination: CHAIR~\cite{rohrbach2018object} and MMHal~\cite{sun2023aligning}. The CHAIR metric quantifies hallucination at both the object level (CHAIR$_\text{I}$) and sentence level (CHAIR$_\text{S}$) as follows: \(
\text{CHAIR}_\text{I} = \frac{|\{\text{hallucinated objects}\}|}{|\{\text{all mentioned objects}\}|}, \quad
\text{CHAIR}_\text{S} = \frac{|\{\text{captions with hallucinated objects}\}|}{|\{\text{all captions}\}|}
\)


In addition, we use MMHal~\cite{sun2023aligning}, a multimodal hallucination benchmark that evaluates object-level consistency using a fine-grained image-text alignment model. Table~\ref{tab:hallucination} reports the hallucination results across different inference-time decoding strategies. Our two-stage value-guided search achieves significant reductions in both CHAIR and MMHal hallucination rates, outperforming all baselines and VisVM-guided search.


\vspace{-5pt}
\begin{table}[h]
  \centering
    \caption{Comparison of visual hallucination and inference efficiency across decoding methods on the COCO Val2017 dataset. CHAIR and MMHal assess hallucination quality ($\uparrow$ $/$ $\downarrow$ indicate better performance), and Avg. Approx Time reports the average inference time per sample in seconds. Top-performing results are highlighted in \textbf{bold}.}

  \label{tab:hallucination}
  \begin{large}
    \resizebox{0.75\textwidth}{!}{%
      \begin{tabular}{l|cccc|c}
        \toprule
        \textbf{Method} 
          & \textbf{CHAIR}$_\text{S}$ $\downarrow$ 
          & \textbf{CHAIR}$_\text{I}$ $\downarrow$ 
          & \textbf{MMHal} $\uparrow$ 
          & \textbf{MMHal Rate} $\downarrow$ 
          & \textbf{Avg. Approx Time} \\
        \midrule
        Greedy Decoding        & 32.4 & 5.9 & 2.94 & 0.52 & \textbf{62 s} \\
        BoN                    & 27.1 & 5.2 & 3.06 & 0.45 & 668 s \\
        CLIP-Guided            & 28.4 & 5.5 & 2.96 & 0.49 & 286 s \\
        VisVM-Guided           & 26.2 & 4.6 & 3.30 & 0.39 & 462 s \\
        \rowcolor{highlight}
        \textbf{ViMaR (Our)} 
                          & \textbf{23.1} & \textbf{4.1} & \textbf{3.75} & \textbf{0.35} & 108 s \\
        \bottomrule
      \end{tabular}%
    }
    \vspace{-5 pt}
  \end{large}
\end{table}

These results demonstrate that our method effectively reduces visual hallucinations during caption generation. Notably, even though our method operates under a smaller decoding budget compared to methods like Visvm, it still yields superior performance. This highlights the efficacy of our localized refinement strategy, which selectively targets visually ambiguous segments for re-generation. Our improvements align with the design of the underlying value model, which is trained to predict long-term rewards using TD learning. By scoring candidate continuations based on their expected future quality—including grounding fidelity—our model promotes selections that reduce hallucination over the entire sequence.

\vspace{-5pt}
\begin{table}[ht]
\centering
\caption{
Evaluation of ViMaR-guided decoding on visual comprehension benchmarks for both LLaVA-Mistral and LLaVA-OneVision-Qwen models. Our two-stage inference framework consistently improves performance across all evaluated tasks, highlighting its effectiveness in enhancing output fidelity and visual grounding. Compared to the base models, ViMaR yields consistent gains, with an average improvement of 15.87\% computed across all evaluation benchmarks, including normalized variants of CHAIRs, CHAIRi, MMHal (normalized as 100--CHAIRs, 10--CHAIRi, and 1--MMHal), and others.
}
\label{tab:performance}
\resizebox{\textwidth}{!}{%

  \begin{tabular}{
  l|l| 
  S[table-format=2.1] S[table-format=2.1] S[table-format=2.1]
  S[table-format=2.1] S[table-format=2.1] S[table-format=2.1]
  S[table-format=2.1]| 
  S[table-format=2.1] S[table-format=2.1] S[table-format=1.2]
  S[table-format=1.2]| 
  S[table-format=2.1, print-implicit-plus = true]
  }
  
\toprule
  & & 
  \multicolumn{7}{c}{Visual Comprehension Benchmark} & 
  \multicolumn{4}{c}{Hallucination Benchmark} & 
  \\
\cmidrule(lr){3-9}\cmidrule(lr){10-13}
Base & SFT Data Source 

& \rotatebox{90}{MM-Vet} $\uparrow$
& \rotatebox{90}{MMBench\,} $\uparrow$
& \rotatebox{90}{MMMU\,} $\uparrow$ 
& \rotatebox{90}{MathVista\,} $\uparrow$ 
& \rotatebox{90}{CVBench\,} $\uparrow$ 
& \rotatebox{90}{LLAVA-W\,} $\uparrow$ 
& \rotatebox{90}{MMStar\,} $\uparrow$ 
& \rotatebox{90}{CHAIRs\,} $\downarrow$ 
& \rotatebox{90}{CHAIRi\,} $\downarrow$ 
& \rotatebox{90}{MMHal\,} $\uparrow$ 
& \rotatebox{90}{MMHal rate\,} $\downarrow$ 
& {Avg.}
\\
\midrule
\multirow{6}{*}{\textbf{LLaVA-Next-Mistral-7B}}
  & Original                     & 45.2 & 74.9 & 34.2 & 38.5 & 65.8 & 76.9 & 36.0 & 32.4 & 5.9 & 2.94 & 0.52 &   \text{--}   \\
  & Greedy decoding        & 43.5 & 74.6 & 34.9 & 37.8 & 66.2 & 75.1 & 36.7 & 33.2 & 6.3 & 2.97 & 0.54 &   -1.44\% \\
  & CLIP-BoN (6)           & 42.8 & 76.2 & 35.2 & 39.7 & 63.8 & 74.8 & 35.5 & 29.7 & 5.2 & 3.05 & 0.48 &    2.45\% \\
  & GPT4-BoN (30)         & 47.1 & 76.1 & 35.4 & 40.9 & 67.9 & 77.3 & 36.9 & 30.0 & 5.4 & 3.11 & 0.47 &   +4.82\% \\
  & CLIP-PRM search       & 46.1 & 75.8 & 35.8 & 39.6 & 68.5 & 78.1 & 37.6 & 26.0 & 5.2 & 3.01 & 0.50 &   +5.33\% \\
  & VisVM search         & 48.3 & 76.7 & 36.1 & 42.3 & 69.8 & 78.4 & 38.0 & 22.6 & 4.3 & 3.26 & 0.44 &  +11.08\% \\
  \rowcolor{highlight}
  &    ViMaR      &\textbf{ 49.8} &\textbf{ 78.2} & \textbf{37.4} & \textbf{42.5} & \textbf{70.7} & \textbf{79.9} & \textbf{39.3} & \textbf{20.8} & \textbf{3.9} & \textbf{3.73} & \textbf{0.38} &  \textbf{+15.87\%} \\

\midrule
  \multirow{1}{*}{\textbf{LLaVA-Onevision-Qwen}}
  & Original              & 58.8 & 81.7 & 47.3 & 56.1 & \text{--} & 86.9 & \text{--} & \text{--} & \text{--} & \text{--} & \text{--} &   \text{--}   \\
  \rowcolor{highlight}
  & ViMaR                     & 60.5 & 84.8 & 49.4 & 56.9 & 80.6 & 88.5 & 62.6 & 15.3 & 3.0 & 3.96 & 0.34 &   \text{--}   \\
  
\bottomrule
\end{tabular}
}
\end{table}

\subsection{Self-Training Vision-Language Model}
\label{sec:self_train}

Beyond its utility at inference time, our two-stage value-guided decoding method offers a compelling opportunity for self-training, leveraging high-quality model-generated responses to further enhance the visual reasoning capabilities of vision-language models (VLM). This section investigates whether the captions produced by our method can serve as effective supervision data for instruction tuning.

\paragraph{Training Setup:}
We construct our supervised fine-tuning (SFT) dataset using the same 23,240 \texttt{<image, prompt>} pairs used for training the value model (as detailed in Section~\ref{sec:vm_training}). Applying our two-stage value-guided decoding strategy, we generate a descriptive caption for each pair, resulting in 23,240 \texttt{<image, prompt, response>} triplets for downstream training. All models are fine-tuned starting from the LLaVA-Next-Mistral-7B checkpoint. To ensure a rigorous and consistent comparison, we adopt the same dataset, evaluation metrics, and scoring setup used in the original VisVM paper. Full-parameter fine-tuning is conducted using a learning rate of 1e-6. We directly compare our approach against the following baselines: greedy decoding, CLIP-based Beam-of-N (BoN), CLIP-PRM guided search, and VisVM-guided search. This evaluation allows us to assess the effectiveness of our search method not only at inference time but also as a mechanism for generating high-quality supervision signals that improve the base model’s visual comprehension through self-training.

\paragraph{Evaluation Benchmarks:}
We evaluate our method across two categories: (i) \textbf{Visual comprehension}, using seven established benchmarks, including MM-Vet \cite{yu2023mm}, MMBench \cite{liu2024mmbench}, MMMU \cite{yue2024mmmu}, MathVista \cite{lu2023mathvista}, CVBench \cite{tong2024cambrian}, LLaVA-Wild \cite{liu2024improved}, and MMStar \cite{chen2024we}; and (ii) \textbf{Hallucination analysis}, assessed via CHAIR \cite{rohrbach2018object} and MMHal \cite{sun2023aligning} metrics. These benchmarks collectively measure the accuracy, reasoning, and visual grounding quality of the generated responses.

\paragraph{Visual Comprehension Results:}  
Table~\ref{tab:performance} summarizes the performance of LLaVA-Next-7B after fine-tuning on captions generated by different inference-time search methods. With the exception of the greedy decoding baseline, which shows marginal declines in some tasks, all self-trained models exhibit improved scores on the suite of visual reasoning benchmarks. Notably, our two-stage value-guided search achieves the most pronounced improvements, with an average uplift of 15.87\% relative to the base model and 4.79\% over the VisVM. These gains markedly outperform those achieved by Best-of-N, CLIP-PRM, and VisVM search, underscoring the exceptional effectiveness of our generated captions as high-quality supervision for advancing VLM visual comprehension.

\paragraph{Visual Hallucination Results:} As presented in Table~\ref{tab:performance}, our proposed two-stage value-guided search demonstrates substantial improvements in mitigating visual hallucinations within LLaVA-Next. When evaluated across four hallucination metrics drawn from CHAIR and MMHal benchmarks, our method achieves a relative reduction of 30.87\% in hallucination rate. This clearly surpasses the improvements observed with CLIP-BoN (7.91\%), GPT4o-BoN (7.82\%), CLIP-PRM search (9.46\%), and VisVM search (20.91\%). These outcomes confirm the robustness of our search strategy in generating responses that are not only detailed but also grounded more accurately in visual content.

\paragraph{Cross-Model Generalization:} 
While ViMaR is trained solely using outputs from LLaVA Mistral-7B, we evaluate its performance when applied to the stronger LLaVA-OneVision-Qwen2-7B model. As shown in Table~\ref{tab:performance}, ViMaR-search yields consistent gains across multiple benchmarks—improving average score on every benchmarks. These results demonstrate that our value model and refinement strategy generalize effectively across architectures, highlighting ViMaR’s flexibility and plug-and-play applicability in high-performing VLMs.

\paragraph{Toward Self-Improving Vision-Language Models:}  
The results highlight the potential of our approach as a self-training paradigm for vision-language models. Importantly, the entire pipeline is constructed without the need for external supervision or third-party models: our value model is trained using the CLIP encoder embedded in LLaVA-Next and initialized with its parameters. The supervised fine-tuning data are generated by leveraging our own inference-time search strategy with LLaVA-Next, ensuring that all learning signals originate from the model itself. This closed-loop design sets the foundation for future extensions of self-training in VLMs, enabling continual performance enhancement without additional human annotations or external models.
\vspace{-5pt}


\subsection{Inference Efficiency } 
\label{sec:time_efficiency}
We evaluate the efficiency of our two-stage value-guided search in terms of average inference time per sample and compare it with several existing decoding strategies, as summarized in Table~\ref{tab:hallucination}. While achieving state-of-the-art performance in reducing visual hallucination, ViMaR remains highly efficient—requiring only 108 seconds per sample on average. In contrast, BoN, CLIP-guided and VisVM-guided search incur significantly higher inference costs (668s, 286s, and 462s, respectively) due to exhaustive scoring or step-by-step evaluation at each generation step. Although greedy decoding is the fastest (62s), it performs the worst across hallucination metrics. ViMaR strikes a strong balance between quality and compute, reducing hallucinations substantially while remaining nearly \textbf{2.6$\times$ faster} than CLIP search, \textbf{4.3$\times$ faster} than VisVM, and over \textbf{6$\times$ faster} than BoN, making it well-suited for practical deployment.

\vspace{-5pt}
\section{Observations and Limitations}\label{sec:obs}
\vspace{-5pt}
To better understand the behavioral differences between our decoding strategy and VisVM, we analyze a representative case where both models are tasked with generating captions for the same image and prompt. As illustrated in Figure~\ref{fig:observation}, LLaVA-Next generates three full-sentence candidates. VisVM selects the final sentence based on local scoring, choosing the third candidate with the highest immediate score (2.285). In contrast, ViMaR evaluates each candidate in the context of the full generated caption from the first stage—considering its broader contribution to overall caption quality. As a result, our model selects the first candidate, which, despite a lower local score, yields the highest predicted global value (2.342) due to its better grounding and potential to lead to more coherent and accurate follow-up content. This example highlights how ViMaR's objective function produces more discriminative and globally aligned scores (e.g., 2.342 vs.\ 1.638), in contrast to VisVM’s closely clustered local scores (e.g., 2.254, 2.216, 2.285), which limit its ability to distinguish high-quality candidates and occasionally result in hallucinated or less grounded outputs.

The lower portion of Figure~\ref{fig:observation} displays the full captions produced following these selections. Our two-stage value-guided approach produces descriptions that are richer in detail and better aligned with the image. For example, it correctly preserves nuanced visual elements such as “\textit{left is holding black and white checkered umbrella with a blue backpack},” while avoiding common hallucinations—such as misattributing visual attributes to the wrong individual. In addition, we visualize the image-text cross-attention maps corresponding to the two selection paths. The attention map from ViMaR shows broader and more balanced coverage of the scene, reflecting the model’s ability to incorporate global context and peripheral visual details. 
Overall, this case study highlights the core distinction between the two approaches: while VisVM performs local step-by-step selection based on immediate reward estimates, our two-stage method first analyzes entire captions to select the most globally coherent candidate, followed by targeted refinement of under-grounded segments. This global-to-local strategy leads to more informed decisions and ultimately more accurate, grounded, and comprehensive descriptions.


\vspace{-5pt}
\section{Conclusion}
\vspace{-5pt}

We introduced \textbf{ViMaR}, a two-stage value-guided inference framework designed to enhance both the efficiency and factual accuracy of VLMs decoding. By integrating a temporal-difference value model with a novel margin-based reward adjustment, ViMaR selectively refined only low-confidence or weakly grounded segments, significantly reducing the computational burden associated with traditional search methods. 
ViMaR achieves substantial improvements in both caption quality and hallucination mitigation, while significantly reducing inference time compared to existing value-guided and search-based decoding strategies.
Qualitative and quantitative evaluation demonstrated that ViMaR exhibits strong cross-model generalization: a value model trained solely on LLaVA Mistral-7B effectively guides generation in the more capable LLaVA-OneVision-Qwen2-7B, reinforcing the modularity and scalability of our inference strategy. Furthermore, when captions generated by ViMaR are used for self-training, the underlying models achieve substantial gains across a diverse suite of visual understanding benchmarks. Together, these results position ViMaR as a fast, accurate, and generalizable decoding framework for improving visual language generation at scale.



\small
\bibliographystyle{plain}
\bibliography{ref}

\begin{thebibliography}{10}

\bibitem{alayrac2022flamingo}
Jean-Baptiste Alayrac, Jeff Donahue, Pauline Luc, Antoine Miech, Iain Barr, Yana Hasson, Karel Lenc, Arthur Mensch, Katherine Millican, Malcolm Reynolds, et~al.
\newblock Flamingo: a visual language model for few-shot learning.
\newblock {\em Advances in neural information processing systems}, 35:23716--23736, 2022.

\bibitem{bai2023qwen}
Jinze Bai, Shuai Bai, Shusheng Yang, Shijie Wang, Sinan Tan, Peng Wang, Junyang Lin, Chang Zhou, and Jingren Zhou.
\newblock Qwen-vl: A frontier large vision-language model with versatile abilities.
\newblock {\em arXiv preprint arXiv:2308.12966}, 1(2):3, 2023.

\bibitem{bai2024hallucination}
Zechen Bai, Pichao Wang, Tianjun Xiao, Tong He, Zongbo Han, Zheng Zhang, and Mike~Zheng Shou.
\newblock Hallucination of multimodal large language models: A survey.
\newblock {\em arXiv preprint arXiv:2404.18930}, 2024.

\bibitem{berger2025improving}
Uri Berger, Omri Abend, Lea Frermann, and Gabriel Stanovsky.
\newblock Improving image captioning by mimicking human reformulation feedback at inference-time.
\newblock {\em arXiv preprint arXiv:2501.04513}, 2025.

\bibitem{betker2023improving}
James Betker, Gabriel Goh, Li~Jing, Tim Brooks, Jianfeng Wang, Linjie Li, Long Ouyang, Juntang Zhuang, Joyce Lee, Yufei Guo, et~al.
\newblock Improving image generation with better captions.
\newblock {\em Computer Science. https://cdn. openai. com/papers/dall-e-3. pdf}, 2(3):8, 2023.

\bibitem{beyer2024paligemma}
Lucas Beyer, Andreas Steiner, Andr{\'e}~Susano Pinto, Alexander Kolesnikov, Xiao Wang, Daniel Salz, Maxim Neumann, Ibrahim Alabdulmohsin, Michael Tschannen, Emanuele Bugliarello, et~al.
\newblock Paligemma: A versatile 3b vlm for transfer.
\newblock {\em arXiv preprint arXiv:2407.07726}, 2024.

\bibitem{brown2024large}
Bradley Brown, Jordan Juravsky, Ryan Ehrlich, Ronald Clark, Quoc~V Le, Christopher R{\'e}, and Azalia Mirhoseini.
\newblock Large language monkeys: Scaling inference compute with repeated sampling.
\newblock {\em arXiv preprint arXiv:2407.21787}, 2024.

\bibitem{chakraborty2024transfer}
Souradip Chakraborty, Soumya~Suvra Ghosal, Ming Yin, Dinesh Manocha, Mengdi Wang, Amrit~Singh Bedi, and Furong Huang.
\newblock Transfer q-star: Principled decoding for llm alignment.
\newblock {\em Advances in Neural Information Processing Systems}, 37:101725--101761, 2024.

\bibitem{chen2024sharegpt4v}
Lin Chen, Jinsong Li, Xiaoyi Dong, Pan Zhang, Conghui He, Jiaqi Wang, Feng Zhao, and Dahua Lin.
\newblock Sharegpt4v: Improving large multi-modal models with better captions.
\newblock In {\em European Conference on Computer Vision}, pages 370--387. Springer, 2024.

\bibitem{chen2024we}
Lin Chen, Jinsong Li, Xiaoyi Dong, Pan Zhang, Yuhang Zang, Zehui Chen, Haodong Duan, Jiaqi Wang, Yu~Qiao, Dahua Lin, et~al.
\newblock Are we on the right way for evaluating large vision-language models?
\newblock {\em arXiv preprint arXiv:2403.20330}, 2024.

\bibitem{chen2024internvl}
Zhe Chen, Jiannan Wu, Wenhai Wang, Weijie Su, Guo Chen, Sen Xing, Muyan Zhong, Qinglong Zhang, Xizhou Zhu, Lewei Lu, et~al.
\newblock Internvl: Scaling up vision foundation models and aligning for generic visual-linguistic tasks.
\newblock In {\em Proceedings of the IEEE/CVF conference on computer vision and pattern recognition}, pages 24185--24198, 2024.

\bibitem{chen2023mitigating}
Zhiyang Chen, Yousong Zhu, Yufei Zhan, Zhaowen Li, Chaoyang Zhao, Jinqiao Wang, and Ming Tang.
\newblock Mitigating hallucination in visual language models with visual supervision.
\newblock {\em arXiv preprint arXiv:2311.16479}, 2023.

\bibitem{fei2021partially}
Zhengcong Fei.
\newblock Partially non-autoregressive image captioning.
\newblock In {\em Proceedings of the AAAI Conference on Artificial Intelligence}, volume~35, pages 1309--1316, 2021.

\bibitem{guan2024hallusionbench}
Tianrui Guan, Fuxiao Liu, Xiyang Wu, Ruiqi Xian, Zongxia Li, Xiaoyu Liu, Xijun Wang, Lichang Chen, Furong Huang, Yaser Yacoob, et~al.
\newblock Hallusionbench: an advanced diagnostic suite for entangled language hallucination and visual illusion in large vision-language models.
\newblock In {\em Proceedings of the IEEE/CVF Conference on Computer Vision and Pattern Recognition}, pages 14375--14385, 2024.

\bibitem{guo2023images}
Jiaxian Guo, Junnan Li, Dongxu Li, Anthony Meng~Huat Tiong, Boyang Li, Dacheng Tao, and Steven Hoi.
\newblock From images to textual prompts: Zero-shot visual question answering with frozen large language models.
\newblock In {\em Proceedings of the IEEE/CVF conference on computer vision and pattern recognition}, pages 10867--10877, 2023.

\bibitem{krause2017hierarchical}
Jonathan Krause, Justin Johnson, Ranjay Krishna, and Li~Fei-Fei.
\newblock A hierarchical approach for generating descriptive image paragraphs.
\newblock In {\em Proceedings of the IEEE conference on computer vision and pattern recognition}, pages 317--325, 2017.

\bibitem{li2023blip}
Junnan Li, Dongxu Li, Silvio Savarese, and Steven Hoi.
\newblock Blip-2: Bootstrapping language-image pre-training with frozen image encoders and large language models.
\newblock In {\em International conference on machine learning}, pages 19730--19742. PMLR, 2023.

\bibitem{li2019coco}
Xirong Li, Chaoxi Xu, Xiaoxu Wang, Weiyu Lan, Zhengxiong Jia, Gang Yang, and Jieping Xu.
\newblock Coco-cn for cross-lingual image tagging, captioning, and retrieval.
\newblock {\em IEEE Transactions on Multimedia}, 21(9):2347--2360, 2019.

\bibitem{lightman2023let}
Hunter Lightman, Vineet Kosaraju, Yuri Burda, Harrison Edwards, Bowen Baker, Teddy Lee, Jan Leike, John Schulman, Ilya Sutskever, and Karl Cobbe.
\newblock Let's verify step by step.
\newblock In {\em The Twelfth International Conference on Learning Representations}, 2023.

\bibitem{liu2024deepseek}
Aixin Liu, Bei Feng, Bin Wang, Bingxuan Wang, Bo~Liu, Chenggang Zhao, Chengqi Dengr, Chong Ruan, Damai Dai, Daya Guo, et~al.
\newblock Deepseek-v2: A strong, economical, and efficient mixture-of-experts language model.
\newblock {\em arXiv preprint arXiv:2405.04434}, 2024.

\bibitem{liu2023mitigating}
Fuxiao Liu, Kevin Lin, Linjie Li, Jianfeng Wang, Yaser Yacoob, and Lijuan Wang.
\newblock Mitigating hallucination in large multi-modal models via robust instruction tuning.
\newblock {\em arXiv preprint arXiv:2306.14565}, 2023.

\bibitem{liu2024improved}
Haotian Liu, Chunyuan Li, Yuheng Li, and Yong~Jae Lee.
\newblock Improved baselines with visual instruction tuning.
\newblock In {\em Proceedings of the IEEE/CVF Conference on Computer Vision and Pattern Recognition}, pages 26296--26306, 2024.

\bibitem{liu2024llavanext}
Haotian Liu, Chunyuan Li, Yuheng Li, Bo~Li, Yuanhan Zhang, Sheng Shen, and Yong~Jae Lee.
\newblock Llava-next: Improved reasoning, ocr, and world knowledge, January 2024.

\bibitem{liu2024mmbench}
Yuan Liu, Haodong Duan, Yuanhan Zhang, Bo~Li, Songyang Zhang, Wangbo Zhao, Yike Yuan, Jiaqi Wang, Conghui He, Ziwei Liu, et~al.
\newblock Mmbench: Is your multi-modal model an all-around player?
\newblock In {\em European conference on computer vision}, pages 216--233. Springer, 2024.

\bibitem{lu2023mathvista}
Pan Lu, Hritik Bansal, Tony Xia, Jiacheng Liu, Chunyuan Li, Hannaneh Hajishirzi, Hao Cheng, Kai-Wei Chang, Michel Galley, and Jianfeng Gao.
\newblock Mathvista: Evaluating mathematical reasoning of foundation models in visual contexts.
\newblock {\em arXiv preprint arXiv:2310.02255}, 2023.

\bibitem{luo2024hallucination}
Junliang Luo, Tianyu Li, Di~Wu, Michael Jenkin, Steve Liu, and Gregory Dudek.
\newblock Hallucination detection and hallucination mitigation: An investigation.
\newblock {\em arXiv preprint arXiv:2401.08358}, 2024.

\bibitem{melas2018training}
Luke Melas-Kyriazi, Alexander~M Rush, and George Han.
\newblock Training for diversity in image paragraph captioning.
\newblock In {\em proceedings of the 2018 conference on empirical methods in natural language processing}, pages 757--761, 2018.

\bibitem{openai2024reasoning}
{OpenAI}.
\newblock Learning to reason with llms.
\newblock \url{https://openai.com/index/learning-to-reason-with-llms/}, 2024.
\newblock Accessed: 2025-04-30.

\bibitem{rennie2017self}
Steven~J Rennie, Etienne Marcheret, Youssef Mroueh, Jerret Ross, and Vaibhava Goel.
\newblock Self-critical sequence training for image captioning.
\newblock In {\em Proceedings of the IEEE conference on computer vision and pattern recognition}, pages 7008--7024, 2017.

\bibitem{rohrbach2018object}
Anna Rohrbach, Lisa~Anne Hendricks, Kaylee Burns, Trevor Darrell, and Kate Saenko.
\newblock Object hallucination in image captioning.
\newblock {\em arXiv preprint arXiv:1809.02156}, 2018.

\bibitem{silver2016mastering}
David Silver, Aja Huang, Chris~J Maddison, Arthur Guez, Laurent Sifre, George Van Den~Driessche, Julian Schrittwieser, Ioannis Antonoglou, Veda Panneershelvam, Marc Lanctot, et~al.
\newblock Mastering the game of go with deep neural networks and tree search.
\newblock {\em nature}, 529(7587):484--489, 2016.

\bibitem{snell2024scaling}
Charlie Snell, Jaehoon Lee, Kelvin Xu, and Aviral Kumar.
\newblock Scaling llm test-time compute optimally can be more effective than scaling model parameters.
\newblock {\em arXiv preprint arXiv:2408.03314}, 2024.

\bibitem{song2024good}
Yifan Song, Guoyin Wang, Sujian Li, and Bill~Yuchen Lin.
\newblock The good, the bad, and the greedy: Evaluation of llms should not ignore non-determinism.
\newblock {\em arXiv preprint arXiv:2407.10457}, 2024.

\bibitem{sun2023aligning}
Zhiqing Sun, Sheng Shen, Shengcao Cao, Haotian Liu, Chunyuan Li, Yikang Shen, Chuang Gan, Liang-Yan Gui, Yu-Xiong Wang, Yiming Yang, et~al.
\newblock Aligning large multimodal models with factually augmented rlhf.
\newblock {\em arXiv preprint arXiv:2309.14525}, 2023.

\bibitem{sutton1988learning}
Richard~S Sutton.
\newblock Learning to predict by the methods of temporal differences.
\newblock {\em Machine learning}, 3:9--44, 1988.

\bibitem{tian2024toward}
Ye~Tian, Baolin Peng, Linfeng Song, Lifeng Jin, Dian Yu, Lei Han, Haitao Mi, and Dong Yu.
\newblock Toward self-improvement of llms via imagination, searching, and criticizing.
\newblock {\em Advances in Neural Information Processing Systems}, 37:52723--52748, 2024.

\bibitem{tong2024cambrian}
Peter Tong, Ellis Brown, Penghao Wu, Sanghyun Woo, Adithya Jairam~Vedagiri IYER, Sai~Charitha Akula, Shusheng Yang, Jihan Yang, Manoj Middepogu, Ziteng Wang, et~al.
\newblock Cambrian-1: A fully open, vision-centric exploration of multimodal llms.
\newblock {\em Advances in Neural Information Processing Systems}, 37:87310--87356, 2024.

\bibitem{uesato2022solving}
Jonathan Uesato, Nate Kushman, Ramana Kumar, Francis Song, Noah Siegel, Lisa Wang, Antonia Creswell, Geoffrey Irving, and Irina Higgins.
\newblock Solving math word problems with process-and outcome-based feedback.
\newblock {\em arXiv preprint arXiv:2211.14275}, 2022.

\bibitem{wan2023faithfulness}
David Wan, Mengwen Liu, Kathleen McKeown, Markus Dreyer, and Mohit Bansal.
\newblock Faithfulness-aware decoding strategies for abstractive summarization.
\newblock {\em arXiv preprint arXiv:2303.03278}, 2023.

\bibitem{wang2024litesearch}
Ante Wang, Linfeng Song, Ye~Tian, Baolin Peng, Dian Yu, Haitao Mi, Jinsong Su, and Dong Yu.
\newblock Litesearch: Efficacious tree search for llm.
\newblock {\em arXiv preprint arXiv:2407.00320}, 2024.

\bibitem{wang2024qwen2}
Peng Wang, Shuai Bai, Sinan Tan, Shijie Wang, Zhihao Fan, Jinze Bai, Keqin Chen, Xuejing Liu, Jialin Wang, Wenbin Ge, et~al.
\newblock Qwen2-vl: Enhancing vision-language model's perception of the world at any resolution.
\newblock {\em arXiv preprint arXiv:2409.12191}, 2024.

\bibitem{wang2024cogvlm}
Weihan Wang, Qingsong Lv, Wenmeng Yu, Wenyi Hong, Ji~Qi, Yan Wang, Junhui Ji, Zhuoyi Yang, Lei Zhao, Song XiXuan, et~al.
\newblock Cogvlm: Visual expert for pretrained language models.
\newblock {\em Advances in Neural Information Processing Systems}, 37:121475--121499, 2024.

\bibitem{wang2024mitigating}
Xintong Wang, Jingheng Pan, Liang Ding, and Chris Biemann.
\newblock Mitigating hallucinations in large vision-language models with instruction contrastive decoding.
\newblock {\em arXiv preprint arXiv:2403.18715}, 2024.

\bibitem{wang2024scaling}
Xiyao Wang, Zhengyuan Yang, Linjie Li, Hongjin Lu, Yuancheng Xu, Chung-Ching Lin, Kevin Lin, Furong Huang, and Lijuan Wang.
\newblock Scaling inference-time search with vision value model for improved visual comprehension.
\newblock {\em arXiv preprint arXiv:2412.03704}, 2024.

\bibitem{wang2024mementos}
Xiyao Wang, Yuhang Zhou, Xiaoyu Liu, Hongjin Lu, Yuancheng Xu, Feihong He, Jaehong Yoon, Taixi Lu, Gedas Bertasius, Mohit Bansal, et~al.
\newblock Mementos: A comprehensive benchmark for multimodal large language model reasoning over image sequences.
\newblock {\em arXiv preprint arXiv:2401.10529}, 2024.

\bibitem{wu2024deepseek}
Zhiyu Wu, Xiaokang Chen, Zizheng Pan, Xingchao Liu, Wen Liu, Damai Dai, Huazuo Gao, Yiyang Ma, Chengyue Wu, Bingxuan Wang, et~al.
\newblock Deepseek-vl2: Mixture-of-experts vision-language models for advanced multimodal understanding.
\newblock {\em arXiv preprint arXiv:2412.10302}, 2024.

\bibitem{xiong2024llava}
Tianyi Xiong, Xiyao Wang, Dong Guo, Qinghao Ye, Haoqi Fan, Quanquan Gu, Heng Huang, and Chunyuan Li.
\newblock Llava-critic: Learning to evaluate multimodal models.
\newblock {\em arXiv preprint arXiv:2410.02712}, 2024.

\bibitem{yang2024qwen2}
An~Yang, Beichen Zhang, Binyuan Hui, Bofei Gao, Bowen Yu, Chengpeng Li, Dayiheng Liu, Jianhong Tu, Jingren Zhou, Junyang Lin, et~al.
\newblock Qwen2. 5-math technical report: Toward mathematical expert model via self-improvement.
\newblock {\em arXiv preprint arXiv:2409.12122}, 2024.

\bibitem{yu2022coca}
Jiahui Yu, Zirui Wang, Vijay Vasudevan, Legg Yeung, Mojtaba Seyedhosseini, and Yonghui Wu.
\newblock Coca: Contrastive captioners are image-text foundation models.
\newblock {\em arXiv preprint arXiv:2205.01917}, 2022.

\bibitem{yu2023mm}
Weihao Yu, Zhengyuan Yang, Linjie Li, Jianfeng Wang, Kevin Lin, Zicheng Liu, Xinchao Wang, and Lijuan Wang.
\newblock Mm-vet: Evaluating large multimodal models for integrated capabilities.
\newblock {\em arXiv preprint arXiv:2308.02490}, 2023.

\bibitem{yue2024mmmu}
Xiang Yue, Yuansheng Ni, Kai Zhang, Tianyu Zheng, Ruoqi Liu, Ge~Zhang, Samuel Stevens, Dongfu Jiang, Weiming Ren, Yuxuan Sun, et~al.
\newblock Mmmu: A massive multi-discipline multimodal understanding and reasoning benchmark for expert agi.
\newblock In {\em Proceedings of the IEEE/CVF Conference on Computer Vision and Pattern Recognition}, pages 9556--9567, 2024.

\bibitem{zhang2024rest}
Dan Zhang, Sining Zhoubian, Ziniu Hu, Yisong Yue, Yuxiao Dong, and Jie Tang.
\newblock Rest-mcts*: Llm self-training via process reward guided tree search.
\newblock {\em Advances in Neural Information Processing Systems}, 37:64735--64772, 2024.

\bibitem{zhou2024calibrated}
Yiyang Zhou, Zhiyuan Fan, Dongjie Cheng, Sihan Yang, Zhaorun Chen, Chenhang Cui, Xiyao Wang, Yun Li, Linjun Zhang, and Huaxiu Yao.
\newblock Calibrated self-rewarding vision language models.
\newblock {\em arXiv preprint arXiv:2405.14622}, 2024.

\end{thebibliography}


\appendix

\section{Human Evaluation}
\label{apx:human_eval}

This section details the human evaluation process used to compare captions generated by ViMaR against those from four decoding baselines: VisVM, CLIP-PRM, best-of-\(N\) (BoN), and greedy decoding. We conduct a blind pairwise comparison study over a randomly sampled subset of 300 image–prompt pairs from the COCO Train2017 dataset, using detailed prompts from the LLaVA-150k dataset.

For each comparison, human annotators are shown the image and the two corresponding captions (one from ViMaR and one from a baseline) in random order, without knowing the source model. Annotators rate which caption is better using a 3-point scale: -1 (baseline is better), 0 (tie), or +1 (ViMaR is better). We aggregate these scores and compute the win rate as the percentage of instances where ViMaR is rated superior (+1).

As reported in Section~\ref{sec:human_gpt_eval}, ViMaR is preferred in 64.0\%, 65.3\%, 66.0\%, and 69.7\% of comparisons against VisVM, CLIP-PRM, BoN, and greedy decoding, respectively. The detailed win rates are also visualized in Figure~\ref{fig:two_gpt_wins}, which summarizes GPT-4o and human preference comparisons across baselines. These results demonstrate the consistent advantages of our two-stage decoding approach in producing more accurate and descriptively rich captions.

\section{GPT Evaluation}
\label{apx_gpt_eval}
In this section, we leverage GPT-4o as an automated judge to compare captions generated by ViMaR against those from baseline decoding strategies. Using the prompt defined above, GPT-4o selects the preferred caption based on richness, accuracy, harmlessness, creativity, and clarity. This large‑scale automated evaluation complements our human studies and metric-based analyses by providing consistent, fine‑grained judgments on caption quality.

\begin{tcolorbox}[
  title=\bfseries GPT-4o Evaluation Prompt,
  colback=white,
  colframe=black,
  boxrule=0.5pt,
  arc=2pt,
  leftrule=1pt,
  rightrule=1pt,
  toprule=1pt,
  bottomrule=1pt,
  fonttitle=\small\bfseries,
  coltitle=black,
  before=\vspace{1ex},
  after=\vspace{1ex},
  left=1em,
  right=1em,
  top=1em,
  bottom=1em
]
Evaluate the following image captions generated by two vision–language models (VLMs) in response to a given image.\\

\textbf{Criteria for “better” caption:}
\begin{itemize}
  \item \textbf{Richness of Content:} Provide a comprehensive description of objects, actions, colors, and settings.
  \item \textbf{Accuracy:} Reflect only what is visible without adding incorrect information.
  \item \textbf{Harmlessness and Appropriateness:} Avoid harmful, offensive, or unwarranted personal assumptions.
  \item \textbf{Creativity and Elaboration:} Offer imaginative yet accurate elaborations that enrich the scene.
  \item \textbf{Clarity and Coherence:} Present a clear, concise, and well-structured description.
\end{itemize}

After considering these, output exactly one of:
\begin{quote}
\texttt{Response1 is better}\\
\texttt{Response2 is better}\\
\texttt{Tie}
\end{quote}

\textbf{Image:} \{Insert image here\}\\
\textbf{Response1:} \{Caption from Model A\}\\
\textbf{Response2:} \{Caption from Model B\}
\end{tcolorbox}



\section{System Configuration and Training Details}
\label{apx:human_exp_setting}

All experiments were conducted on a single NVIDIA RTX A6000 GPU with 48 GB of VRAM. We utilized mixed-precision training with \texttt{fp16} to optimize memory usage and computational throughput. The training process was launched using the \texttt{accelerate} framework with gradient checkpointing enabled to reduce memory overhead.
The model was fine-tuned on the \texttt{LLAVA} dataset using the provided \texttt{train} and \texttt{test} splits, with a per-device batch size of 16. Training was performed over 4 epochs.
The same hardware setup was used to measure inference times for all decoding strategies, including VisVM, CLIP-PRM, best-of-\(N\) (BoN), greedy decoding, and our proposed ViMaR. All evaluations were conducted under identical conditions and batch sizes to ensure a fair and consistent comparison of both efficiency and performance.

\section{Analysis and Selection of Margin Threshold \texorpdfstring{$\tau$}{tau}}
\label{apx:tau_selection}

To ensure effective reward shaping during value model training, we empirically analyze the distribution of CLIP similarity scores across the full training set to determine a principled value for the margin threshold $\tau$. Our margin-based penalty mechanism is activated when a candidate caption's CLIP similarity score falls below $\tau$, enforcing a negative reward proportional to the gap. The choice of $\tau$ directly governs the aggressiveness of this penalty and thus requires careful calibration.

We compute summary statistics over the entire dataset's CLIP similarity scores, resulting in the following: lowest score = 0.0031, highest = 0.4580, mean = 0.2102. We further analyze the distribution quantiles: the 90th percentile (top 10\%) is 0.2749, the 80th percentile is 0.2544, the 20th percentile is 0.1636, and the 10th percentile is 0.1429. Based on this, we select $\tau = 0.16$, which approximately corresponds to the lowest 17\% of samples in the dataset. This threshold captures a meaningful boundary between well-grounded and underperforming captions, ensuring that only semantically weak generations receive penalization during training.

This percentile-based approach allows us to define $\tau$ in a data-driven, distribution-aware manner, avoiding manual tuning and yielding a stable learning signal. By anchoring the penalty trigger to the empirical distribution, we promote robustness and generalizability of the margin-based reward across diverse datasets and captioning scenarios. The integration of this threshold into our value model's training objective is detailed in Section~\ref{sec:td_training}, where we describe the temporal-difference learning framework and margin-based reward adjustment.

\section{Additional Case Studies}
\label{apx:qualitative}
In this section, we present further qualitative comparisons to illustrate the differences between ViMaR-guided decoding and baseline methods. Figures~\ref{fig:supp1} through~\ref{fig:supp3} showcase a series of representative examples, highlighting how our approach improves caption fidelity, visual grounding, and descriptive richness across diverse scenes. Additional qualitative results are also provided in the supplementary material to further support our findings.

\clearpage
\begin{figure}
    \centering
    \includegraphics[width=1.0\linewidth]{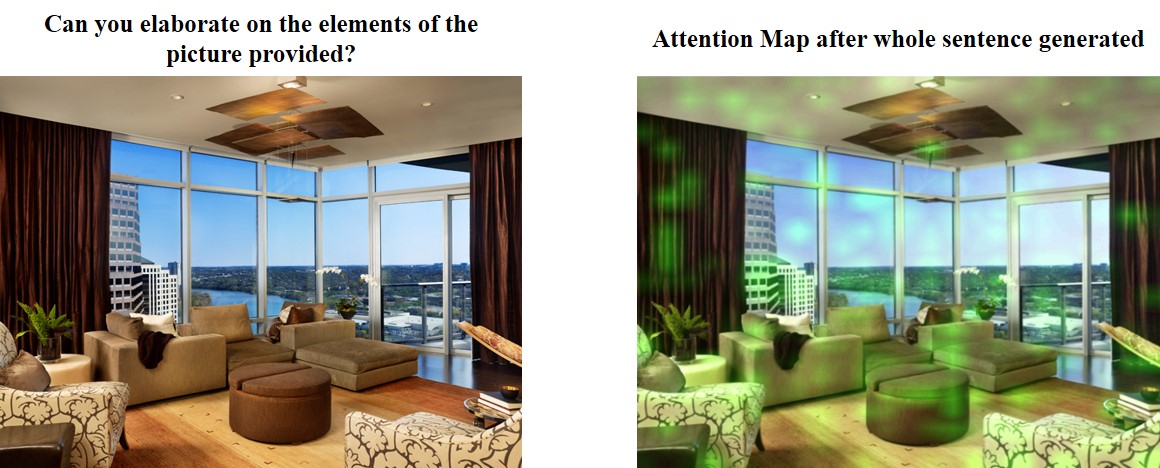}
    \hfill
    \includegraphics[width=1.0\linewidth]{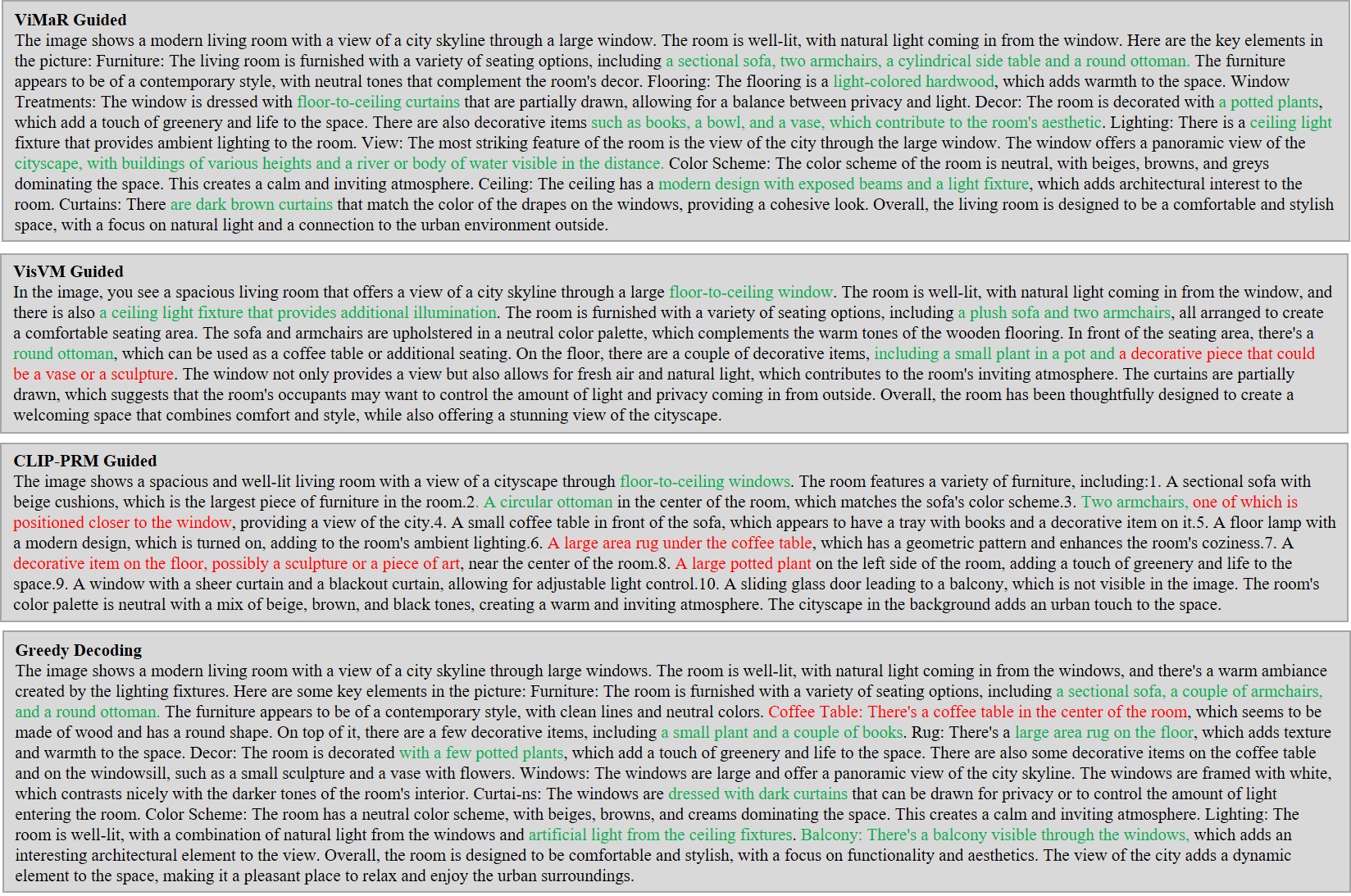}
    \caption{Qualitative comparison example 1}
    \label{fig:supp1}
\end{figure}

\clearpage
\begin{figure}
    \centering
    \includegraphics[width=1.0\linewidth]{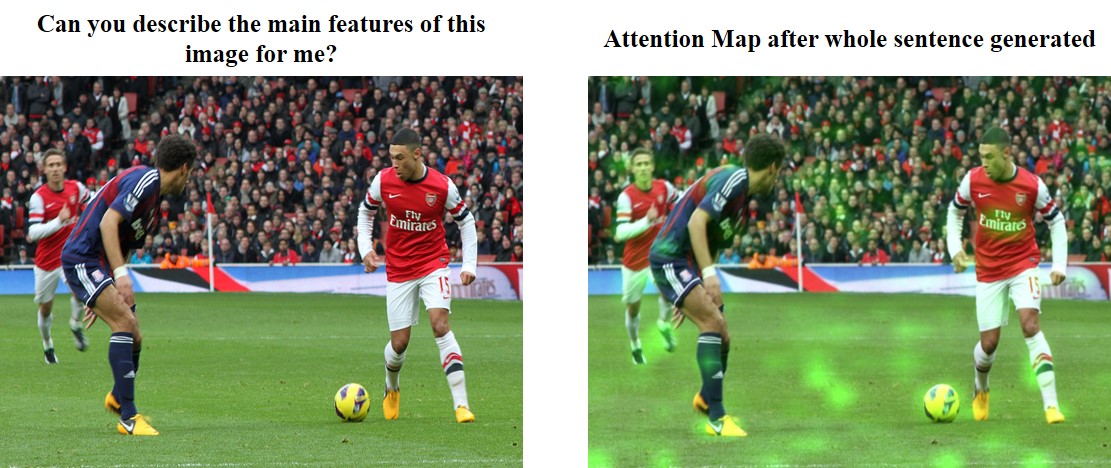}
    \hfill
    \includegraphics[width=1.0\linewidth]{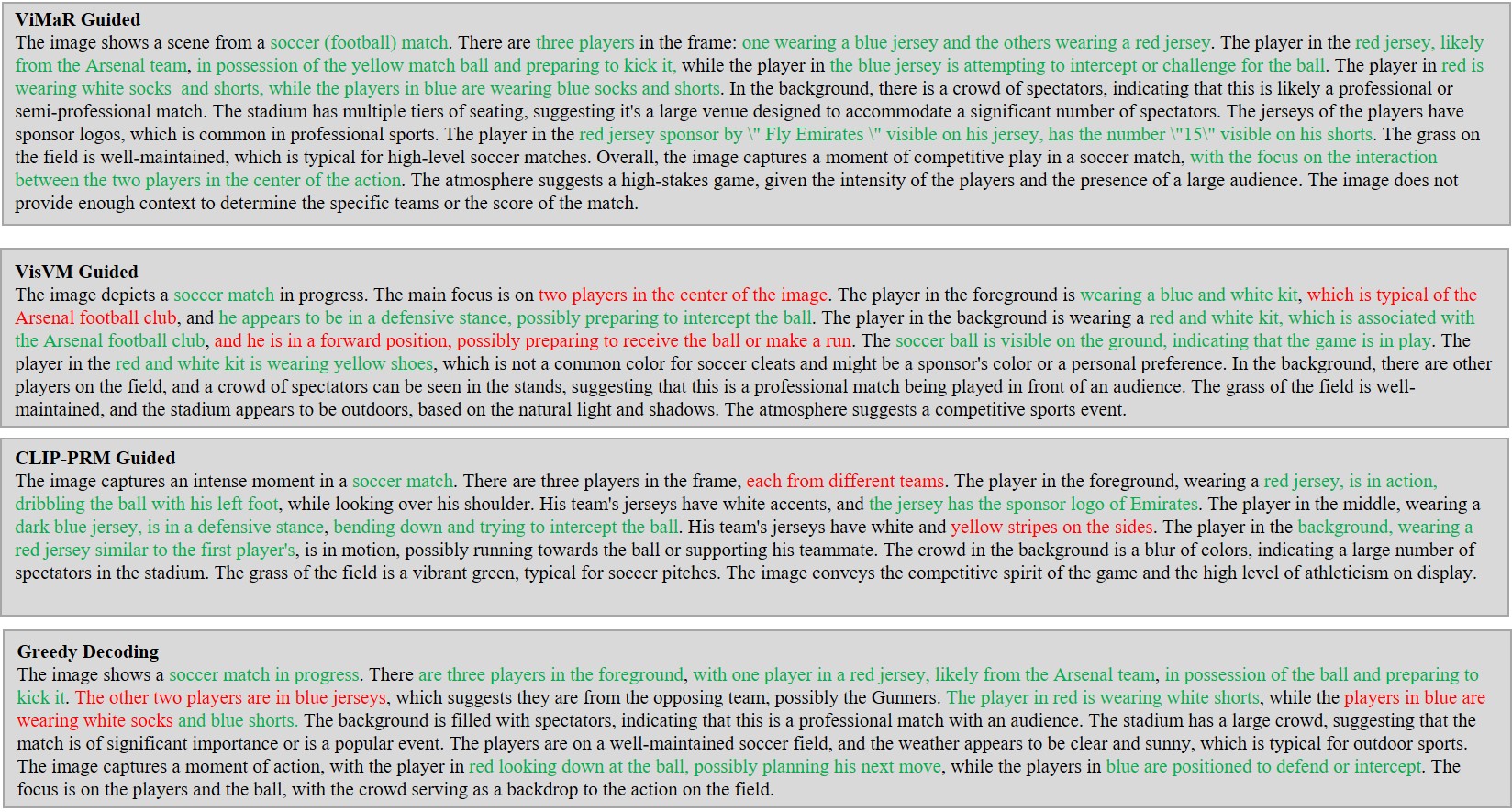}
    \caption{Qualitative comparison example 2}
    \label{fig:supp2}
\end{figure}

\clearpage
\begin{figure}
    \centering
    \includegraphics[width=1.0\linewidth]{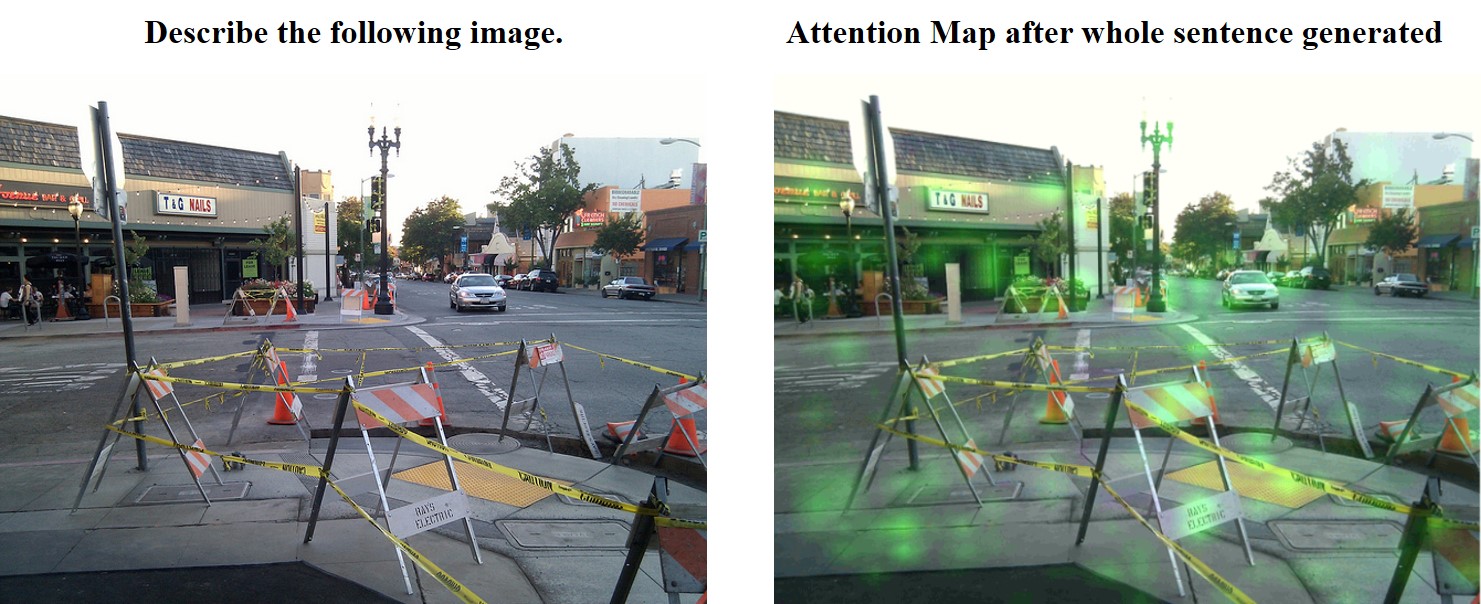}
    \hfill
    \includegraphics[width=1.0\linewidth]{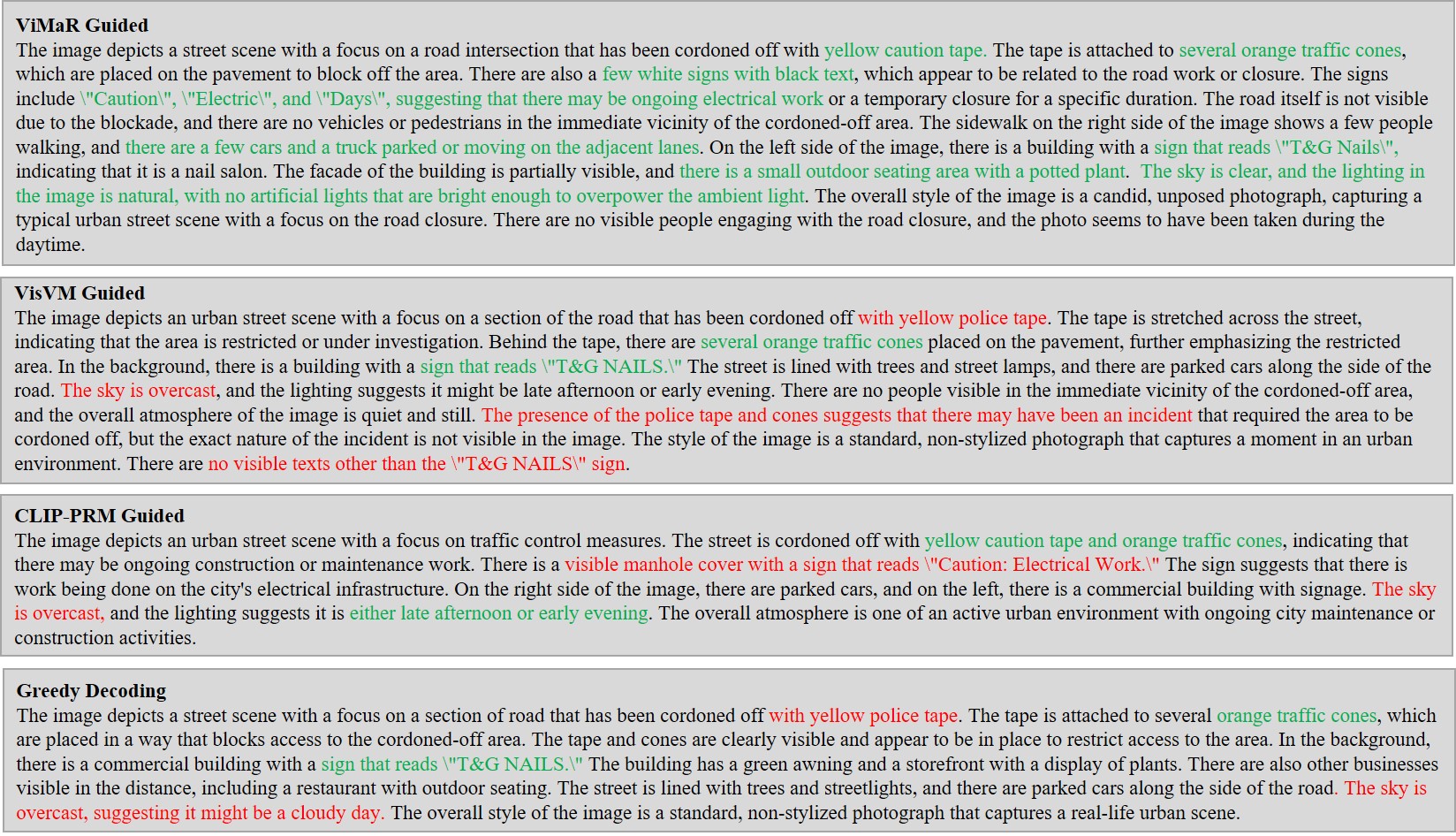}
    \caption{Qualitative comparison example 3}
    \label{fig:supp3}
\end{figure}

 
 
 

 
 
 

\clearpage
\section{Extended Qualitative Results}
\label{supp:extended_qualitative}

\begin{figure}
    \centering
    \includegraphics[width=1.0\linewidth]{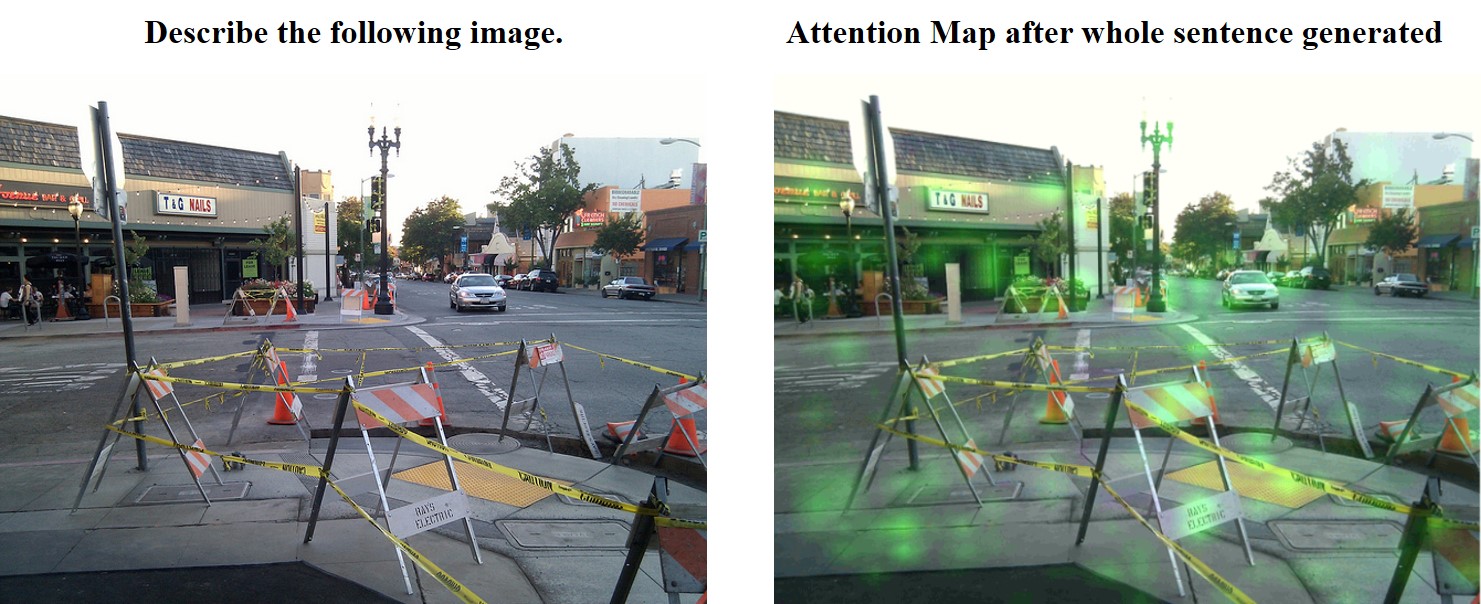}
    \hfill
    \includegraphics[width=1.0\linewidth]{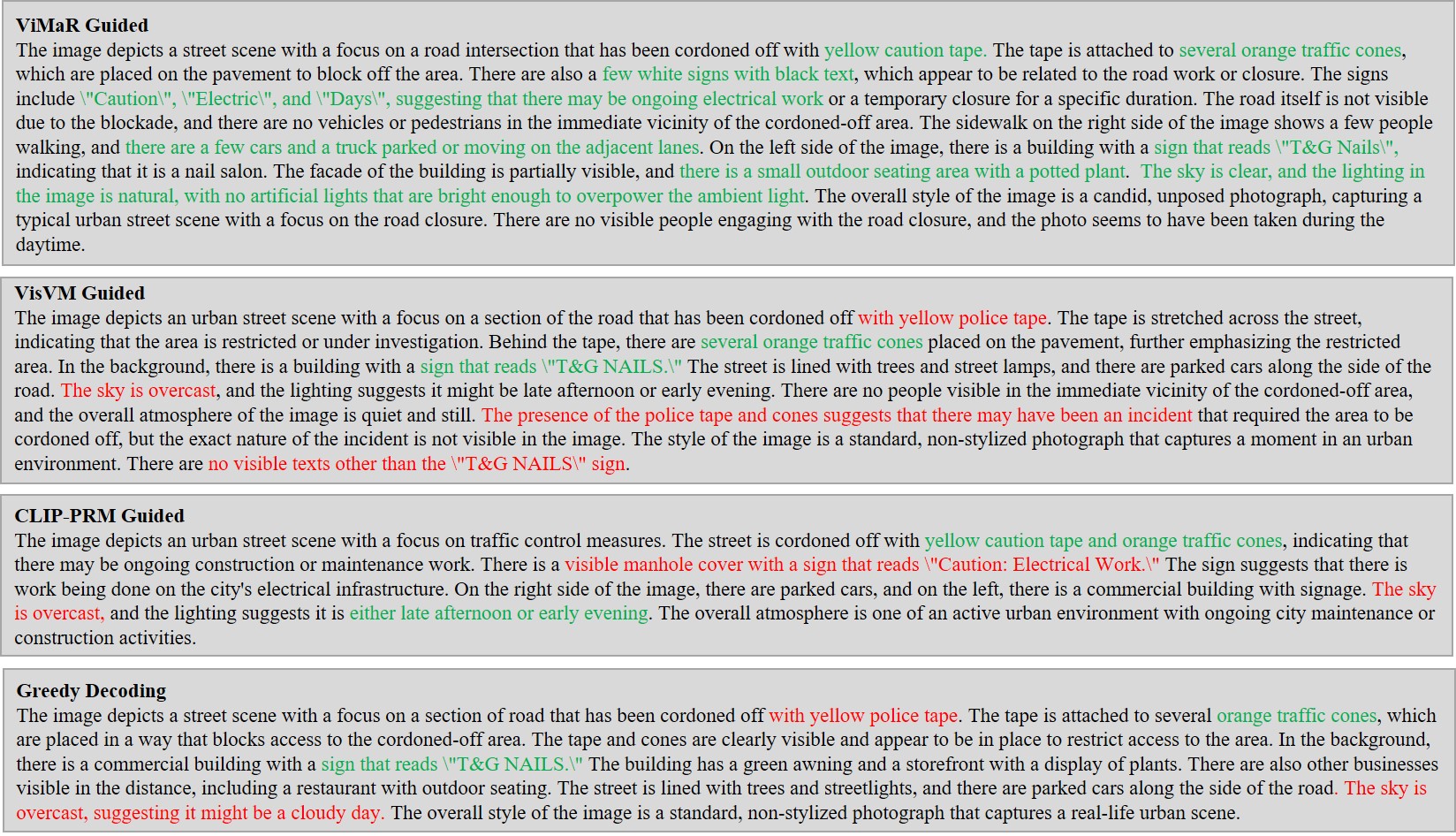}
    \caption{Case study 14: ViMaR and baseline captions.}
    \label{fig:supp14}
\end{figure}

This section provides additional qualitative examples to supplement those presented in main paper and appendix of the main paper. We include a broader set of visualization cases comparing ViMaR-guided decoding with baseline methods across varied scenes and prompt types. Figures~\ref{fig:supp1} through~\ref{fig:supp14} showcase a series of challenging and representative examples that require fine-grained visual understanding and precise language generation. These extended examples further highlight ViMaR’s ability to produce more visually grounded, detailed, and hallucination-resistant captions. Each figure illustrates both the model outputs and the corresponding attention maps to facilitate deeper analysis of decoding behavior.

\clearpage
\begin{figure}
    \centering
    \includegraphics[width=1.0\linewidth]{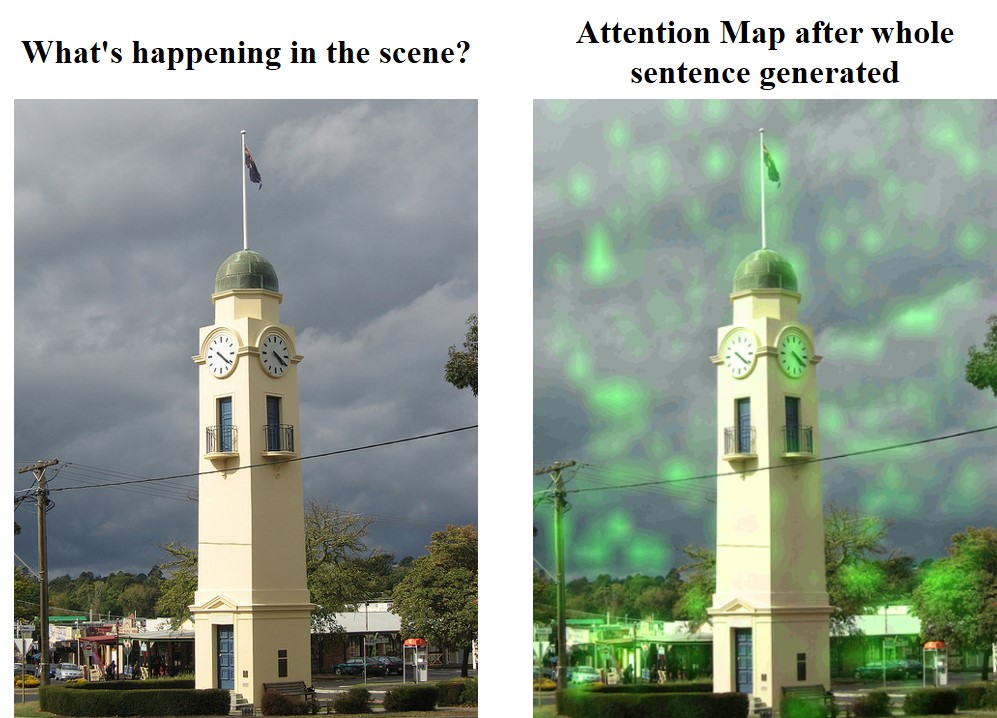}
    \hfill
    \includegraphics[width=1.0\linewidth]{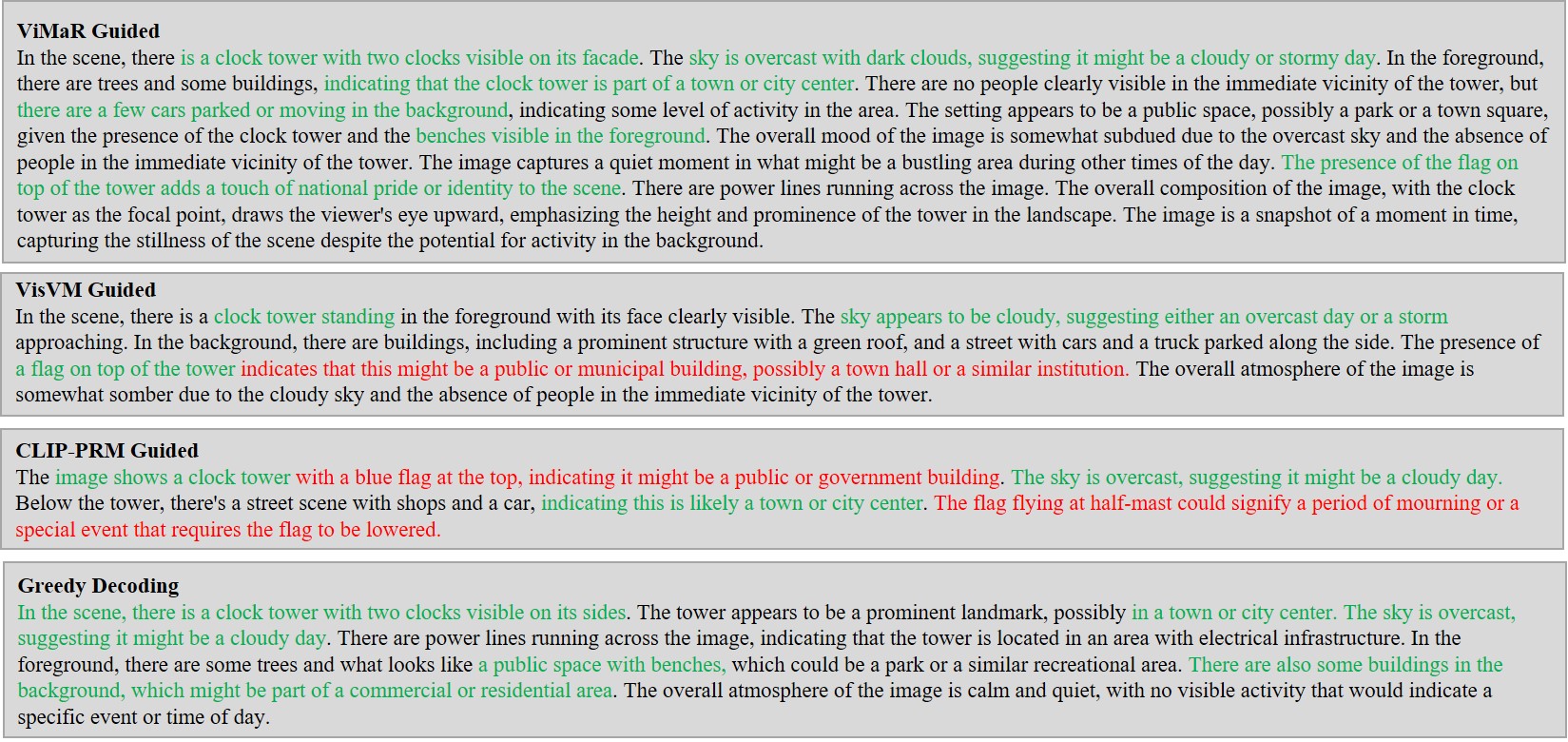}
    \caption{Case study 1: ViMaR and baseline captions.}
    \label{fig:supp1}
\end{figure}

\clearpage
\begin{figure}
    \centering
    \includegraphics[width=1.0\linewidth]{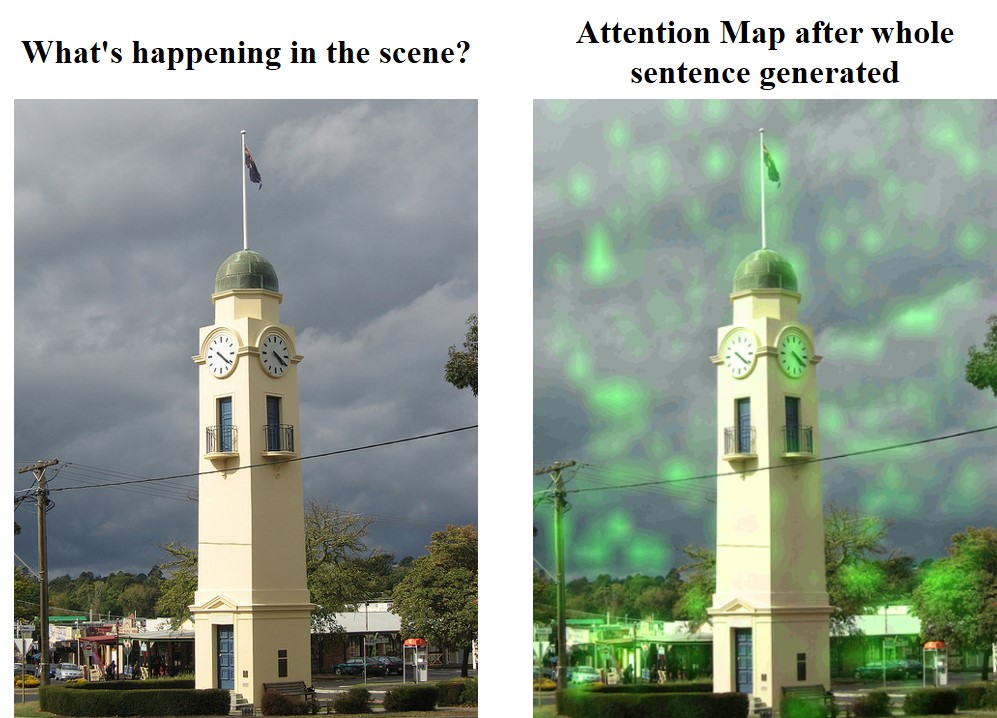}
    \hfill
    \includegraphics[width=1.0\linewidth]{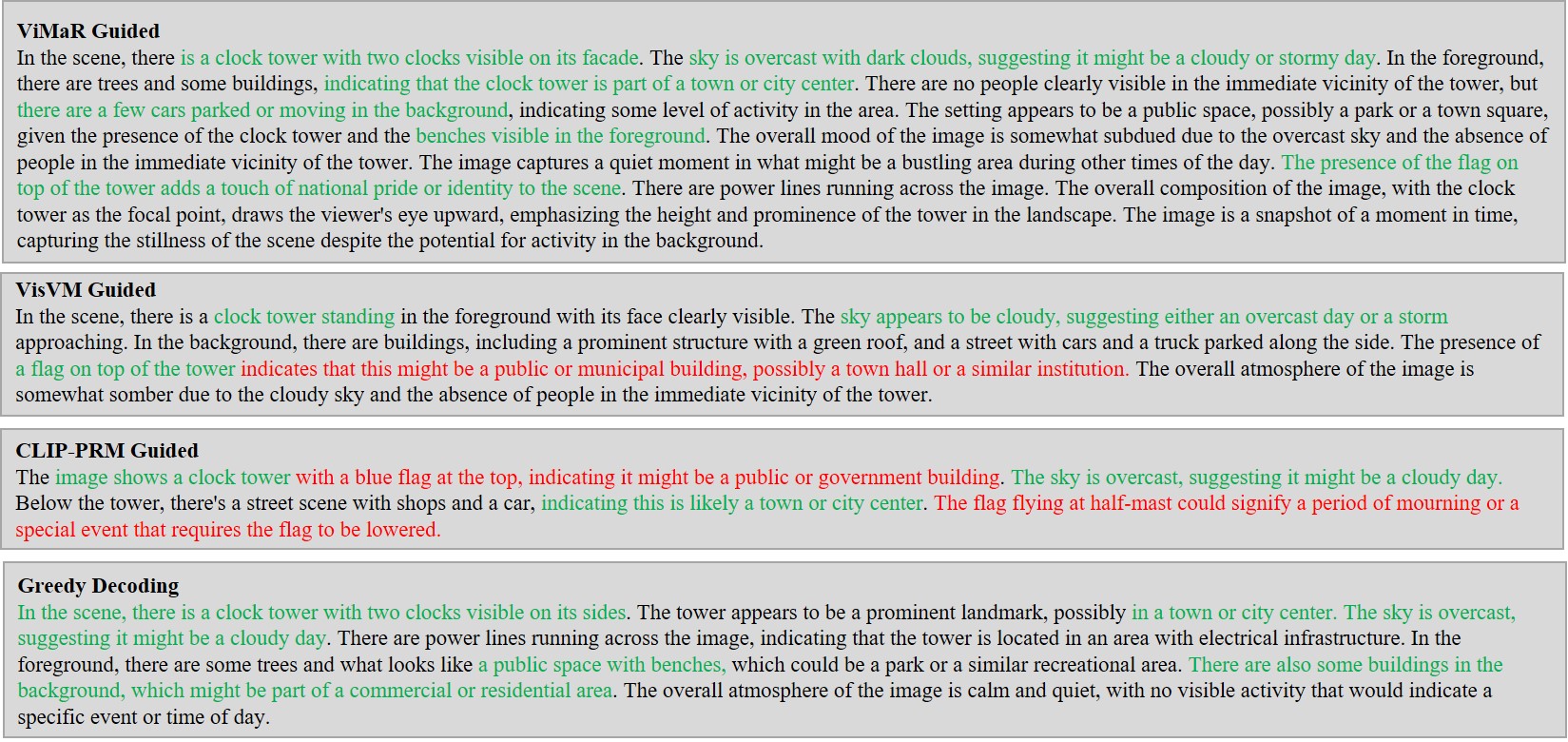}
    \caption{Case study 2: ViMaR and baseline captions.}
    \label{fig:supp2}
\end{figure}

\clearpage
\begin{figure}
    \centering
    \includegraphics[width=1.0\linewidth]{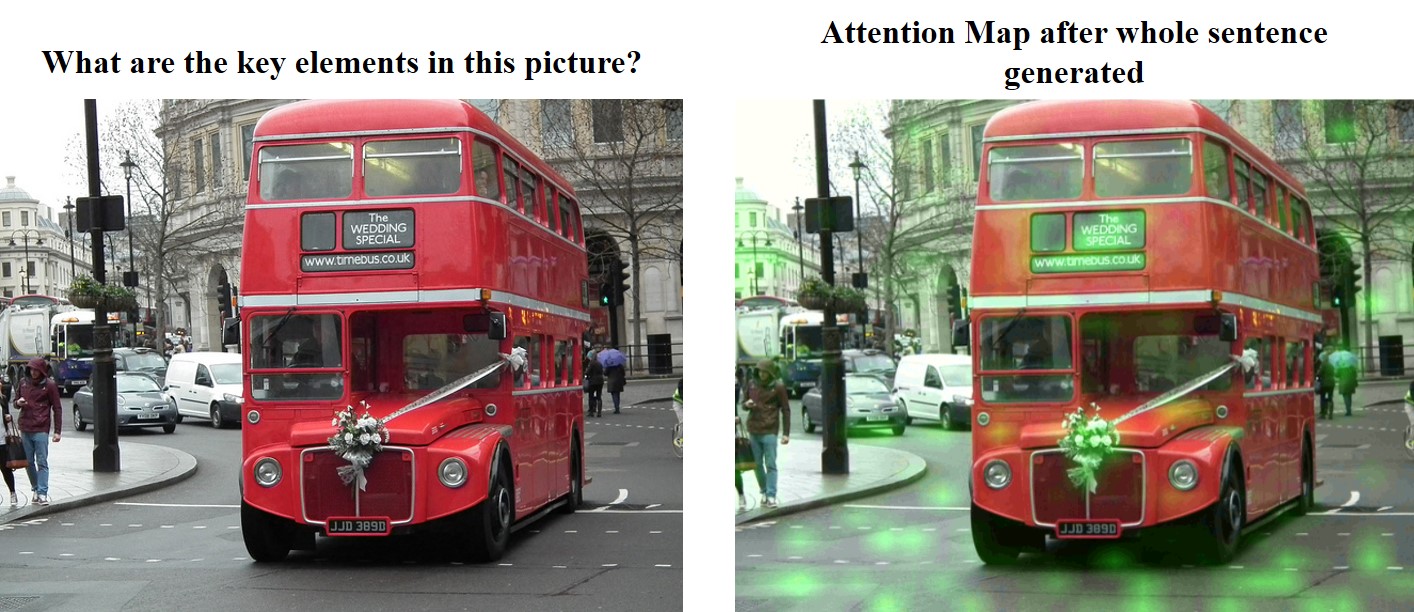}
    \hfill
    \includegraphics[width=1.0\linewidth]{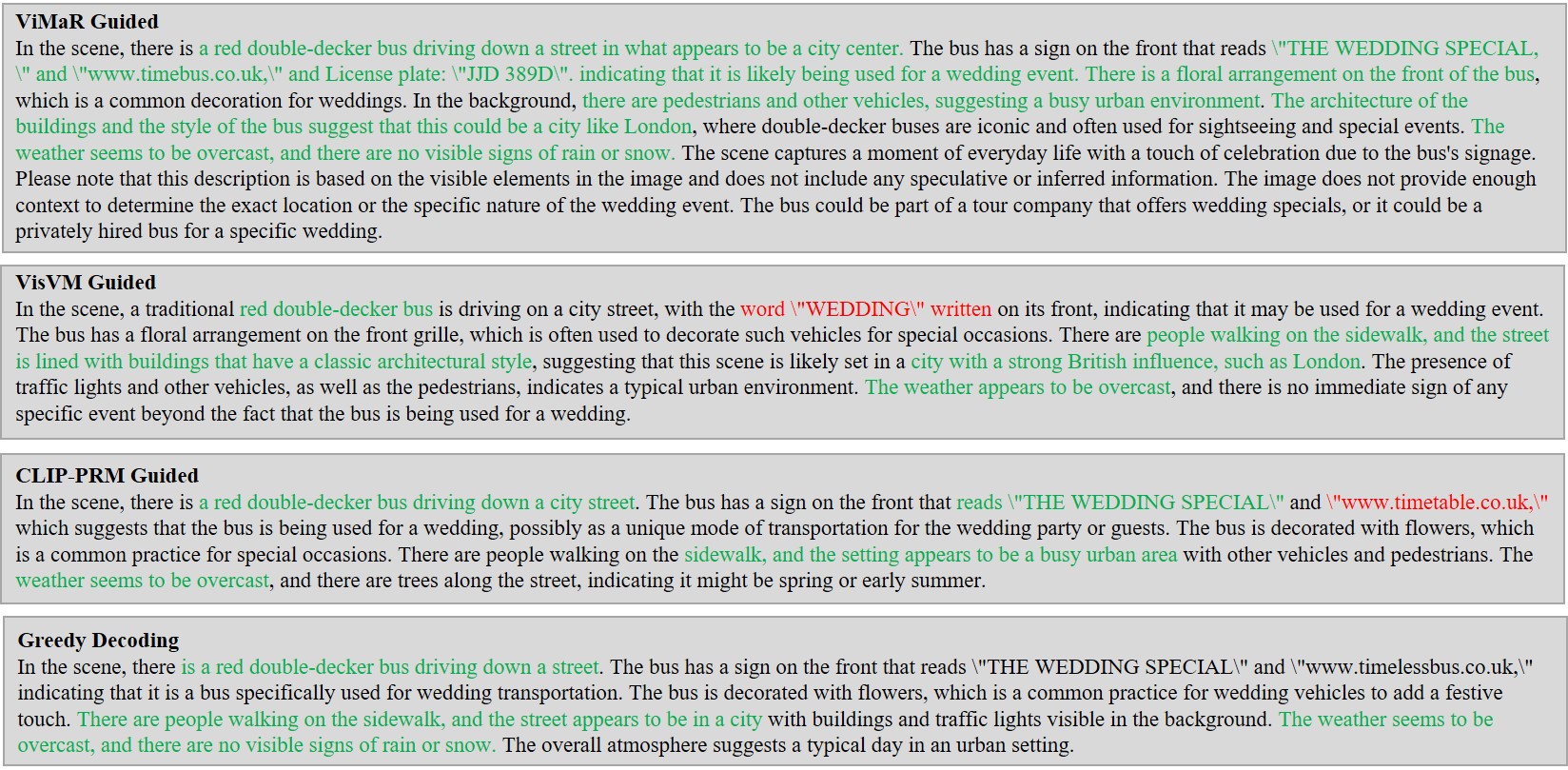}
    \caption{Case study 3: ViMaR and baseline captions.}
    \label{fig:supp3}
\end{figure}

\clearpage
\begin{figure}
    \centering
    \includegraphics[width=1.0\linewidth]{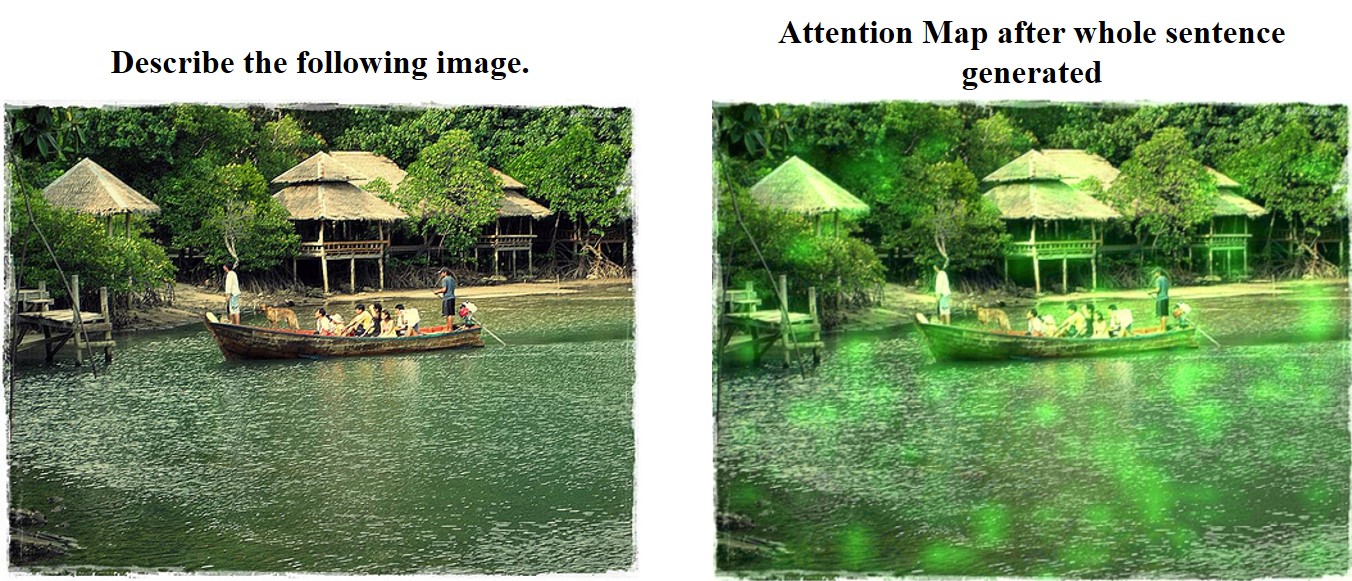}
    \hfill
    \includegraphics[width=1.0\linewidth]{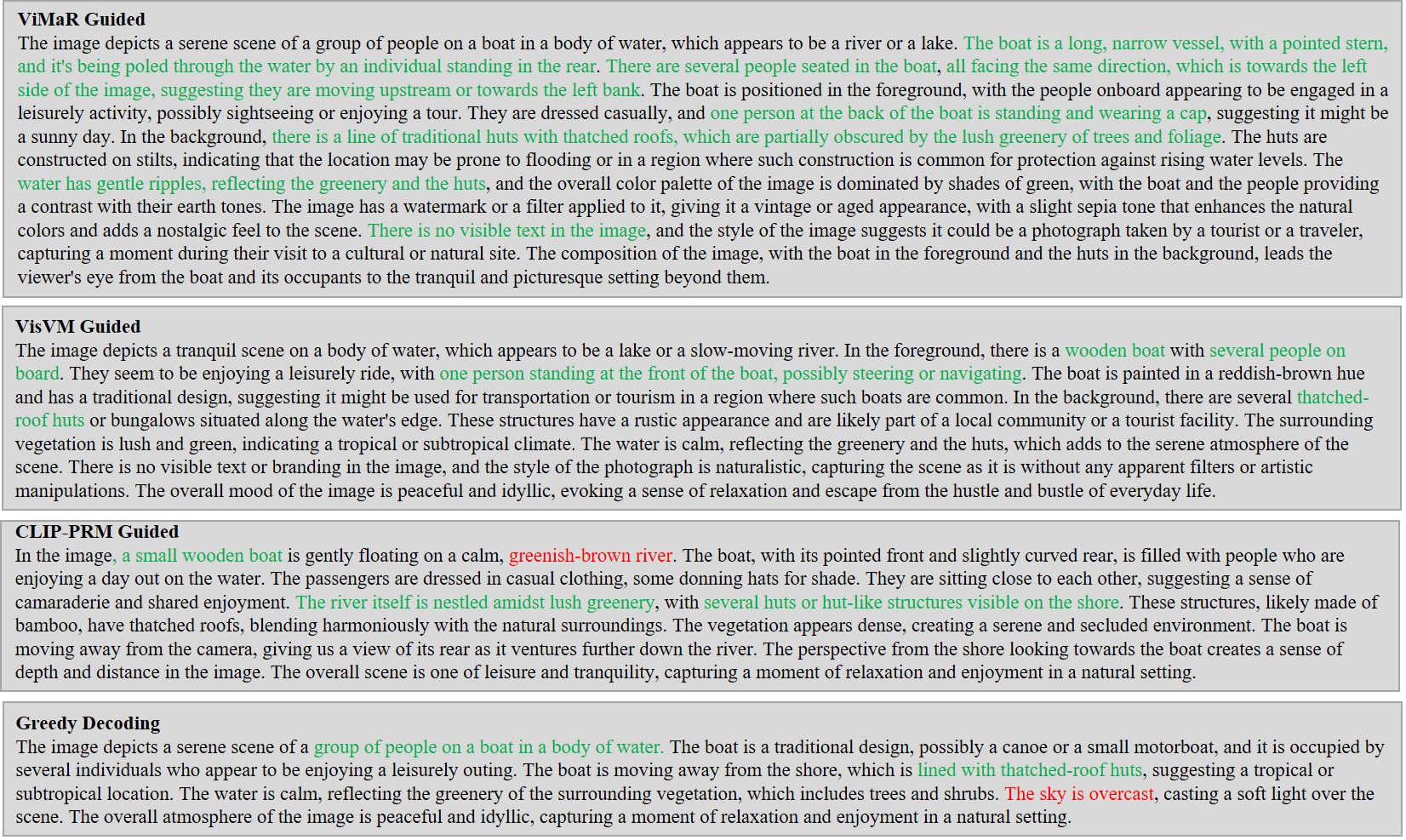}
    \caption{Case study 4: ViMaR and baseline captions.}
    \label{fig:supp4}
\end{figure}

\clearpage
\begin{figure}
    \centering
    \includegraphics[width=1.0\linewidth]{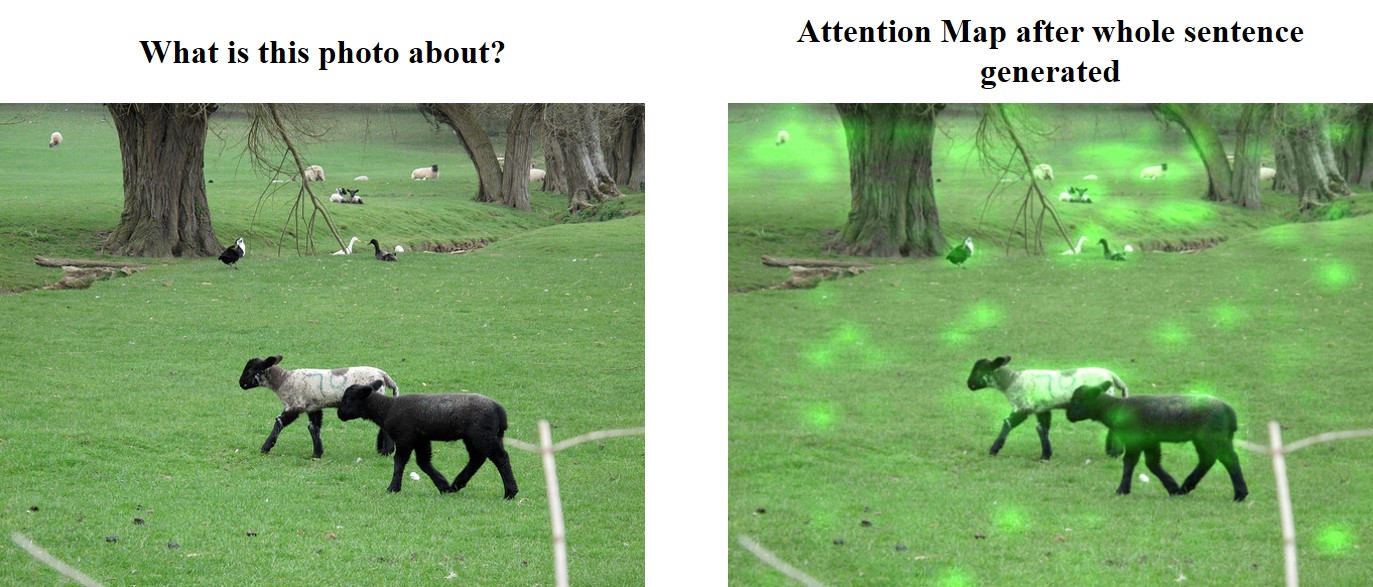}
    \hfill
    \includegraphics[width=1.0\linewidth]{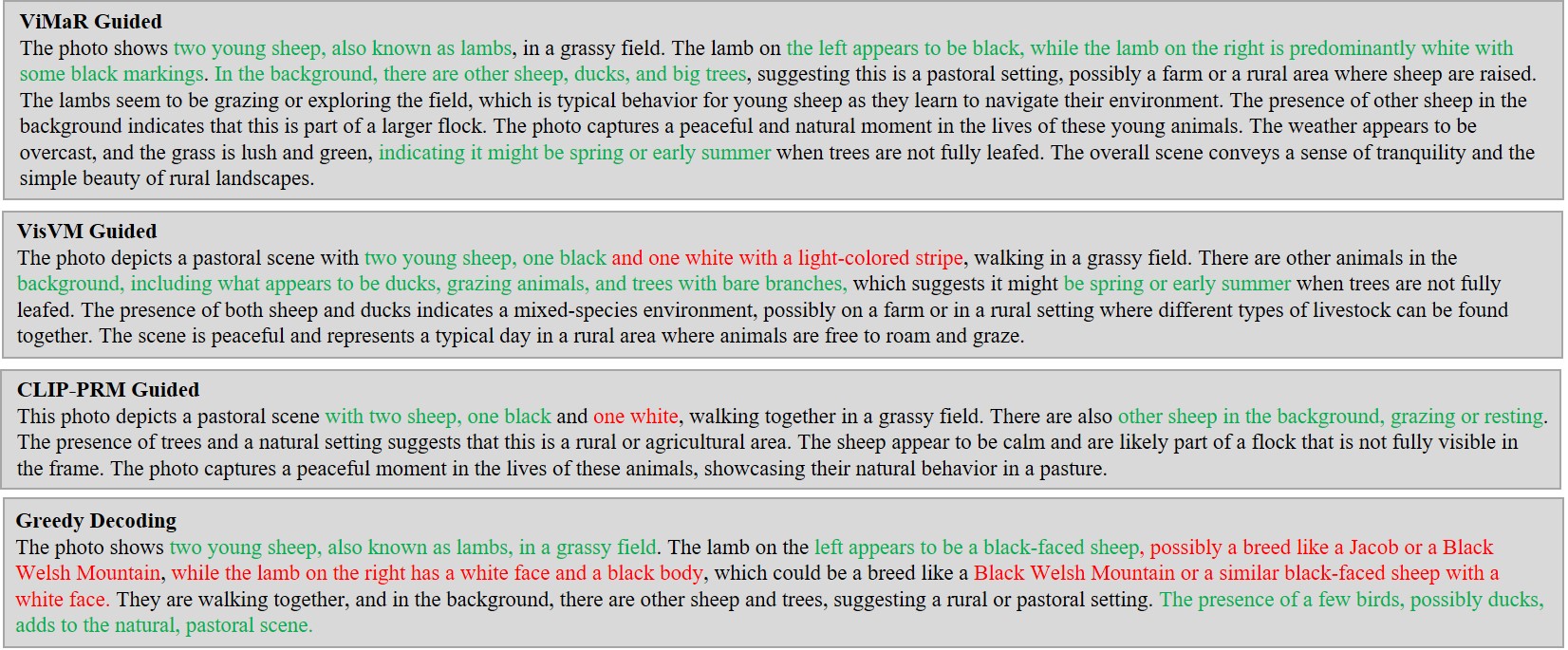}
    \caption{Case study 5: ViMaR and baseline captions.}
    \label{fig:supp5}
\end{figure}

\clearpage
\begin{figure}
    \centering
    \includegraphics[width=1.0\linewidth]{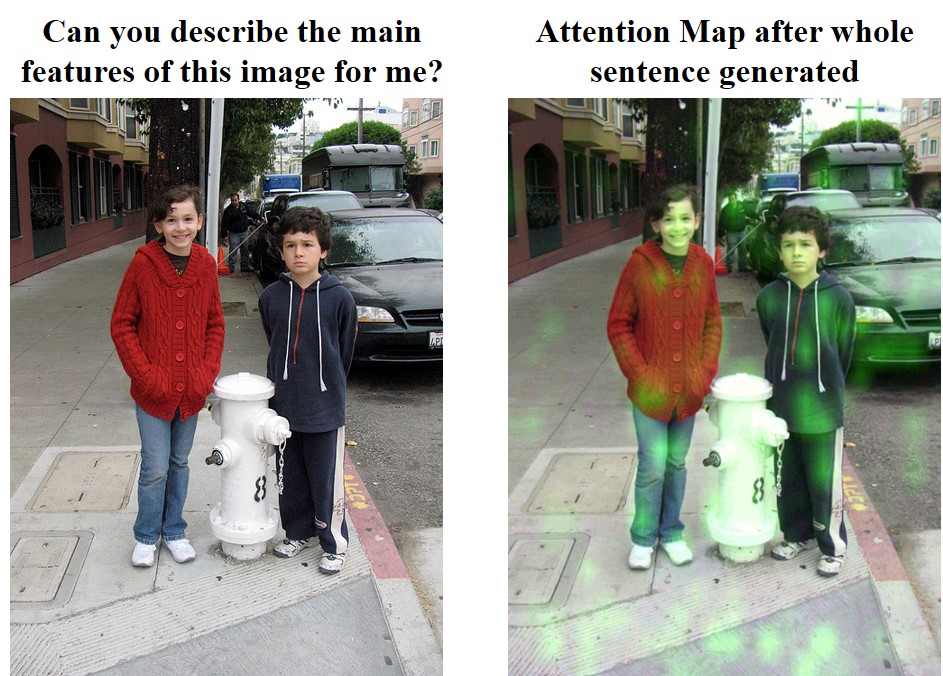}
    \hfill
    \includegraphics[width=1.0\linewidth]{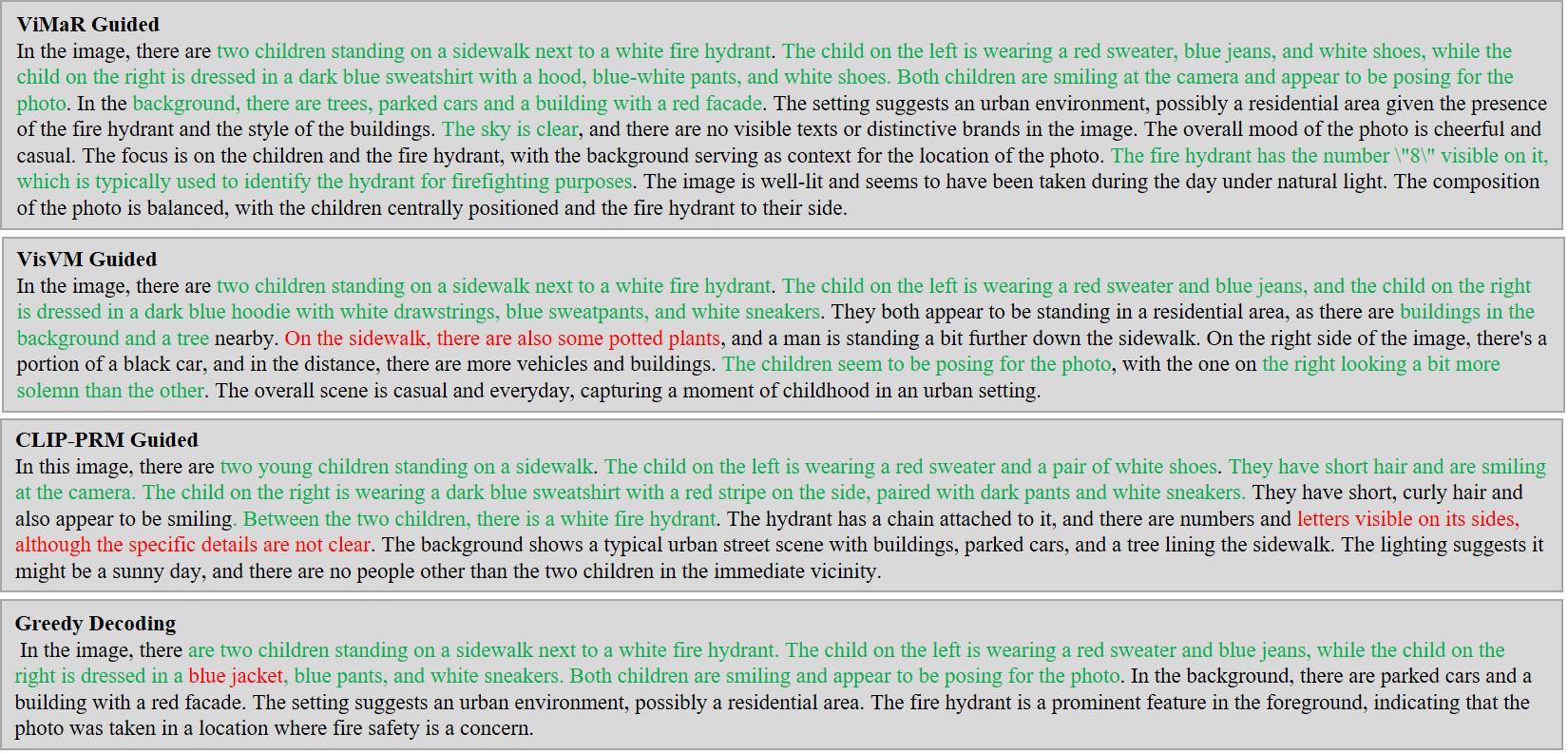}
    \caption{Case study 6: ViMaR and baseline captions.}
    \label{fig:supp6}
\end{figure}

\clearpage
\begin{figure}
    \centering
    \includegraphics[width=1.0\linewidth]{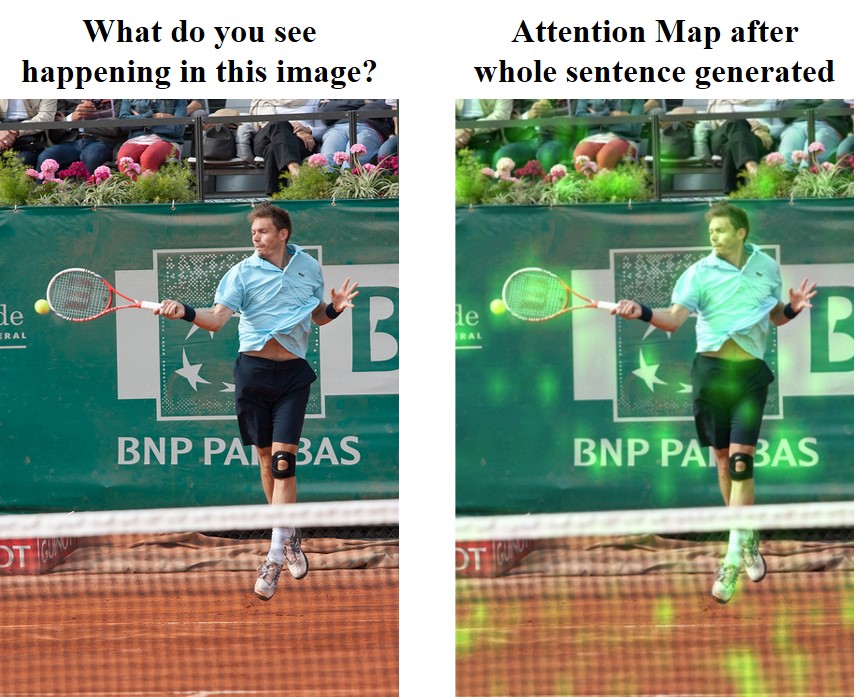}
    \hfill
    \includegraphics[width=1.0\linewidth]{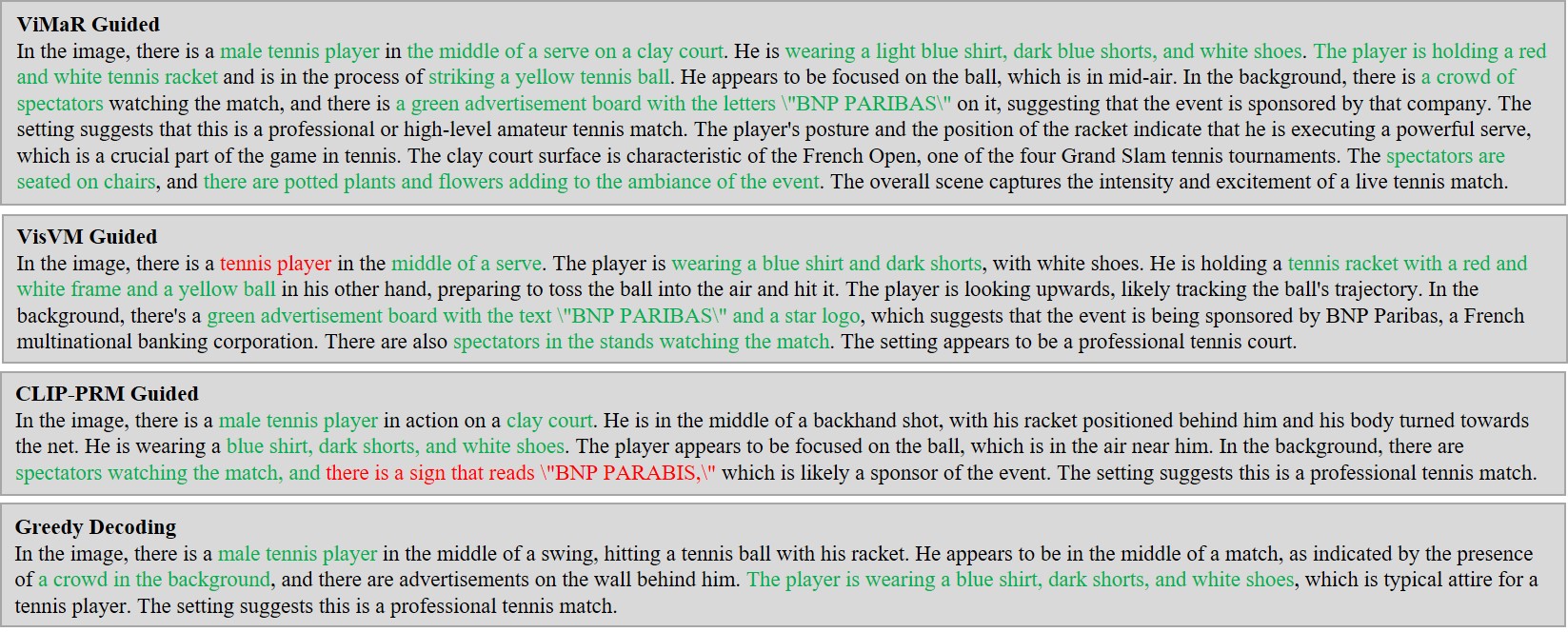}
    \caption{Case study 7: ViMaR and baseline captions.}
    \label{fig:supp7}
\end{figure}

\clearpage
\begin{figure}
    \centering
    \includegraphics[width=1.0\linewidth]{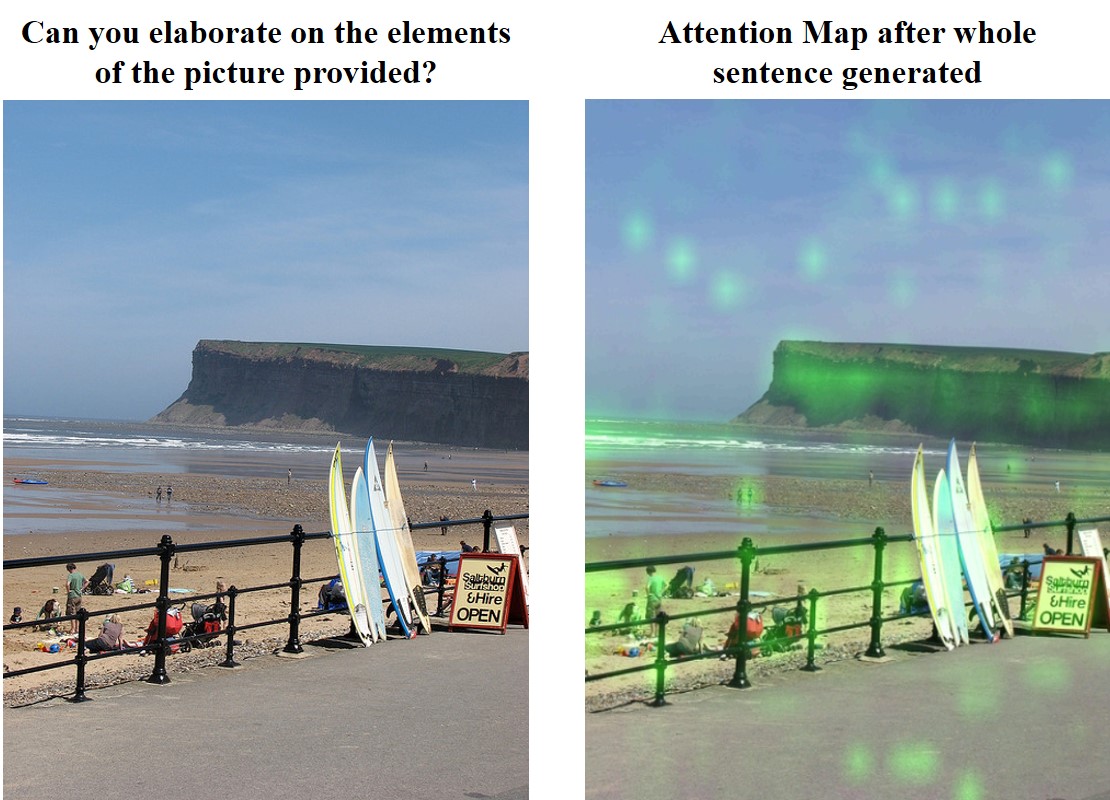}
    \hfill
    \includegraphics[width=1.0\linewidth]{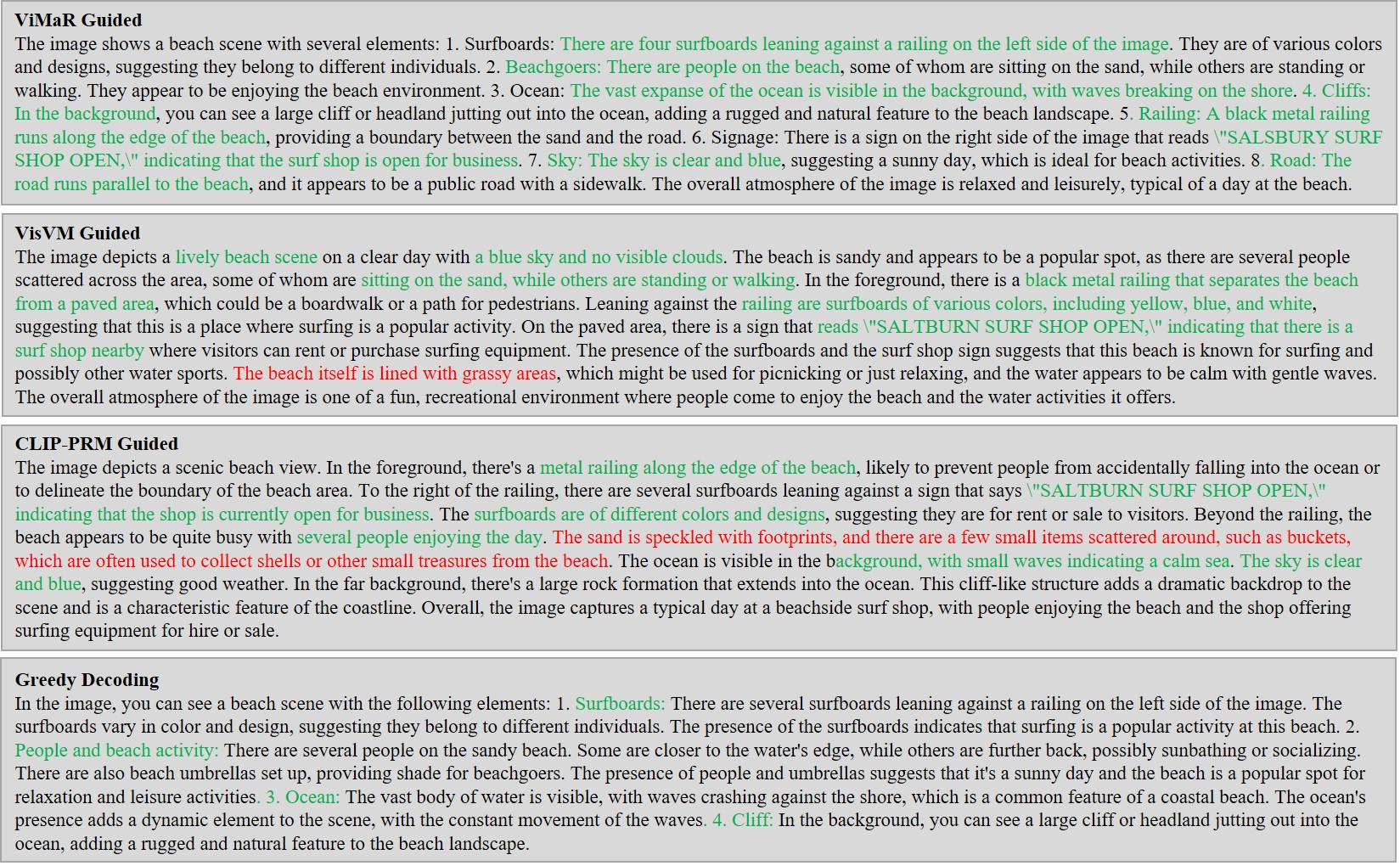}
    \caption{Case study 8: ViMaR and baseline captions.}
    \label{fig:supp8}
\end{figure}

\clearpage
\begin{figure}
    \centering
    \includegraphics[width=1.0\linewidth]{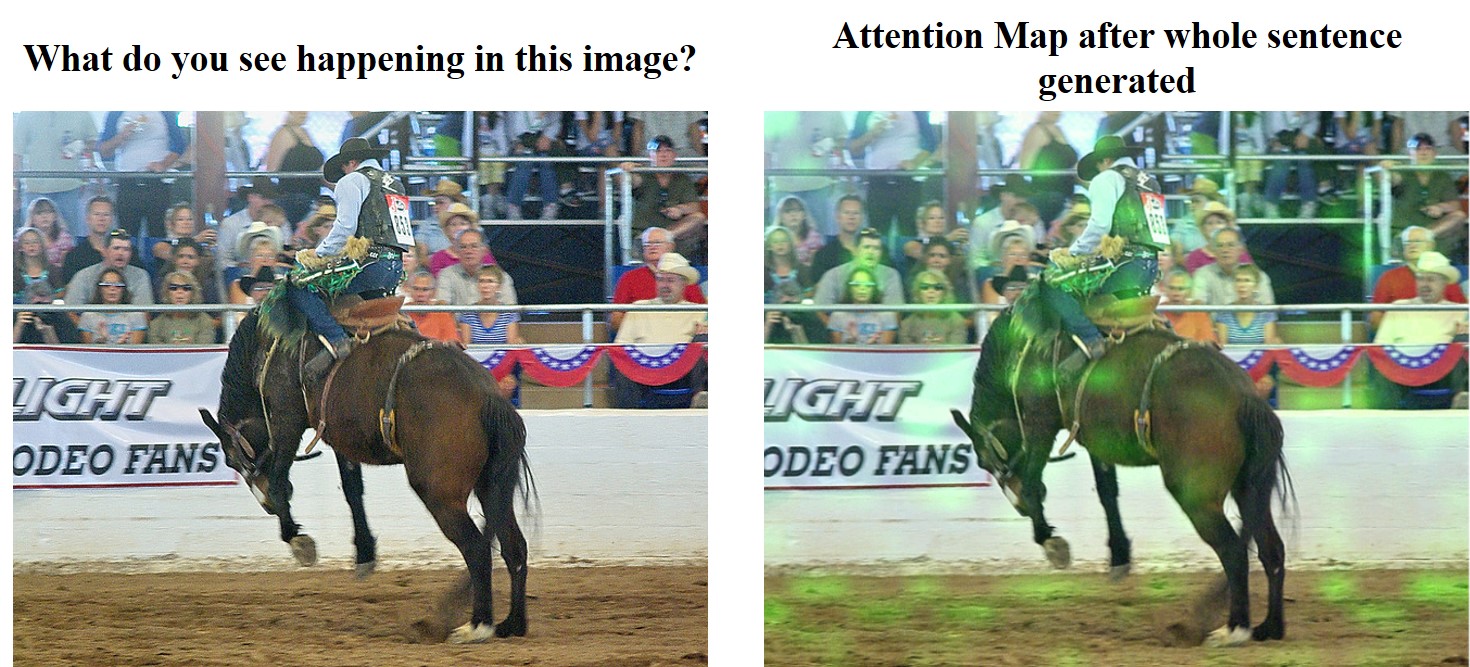}
    \hfill
    \includegraphics[width=1.0\linewidth]{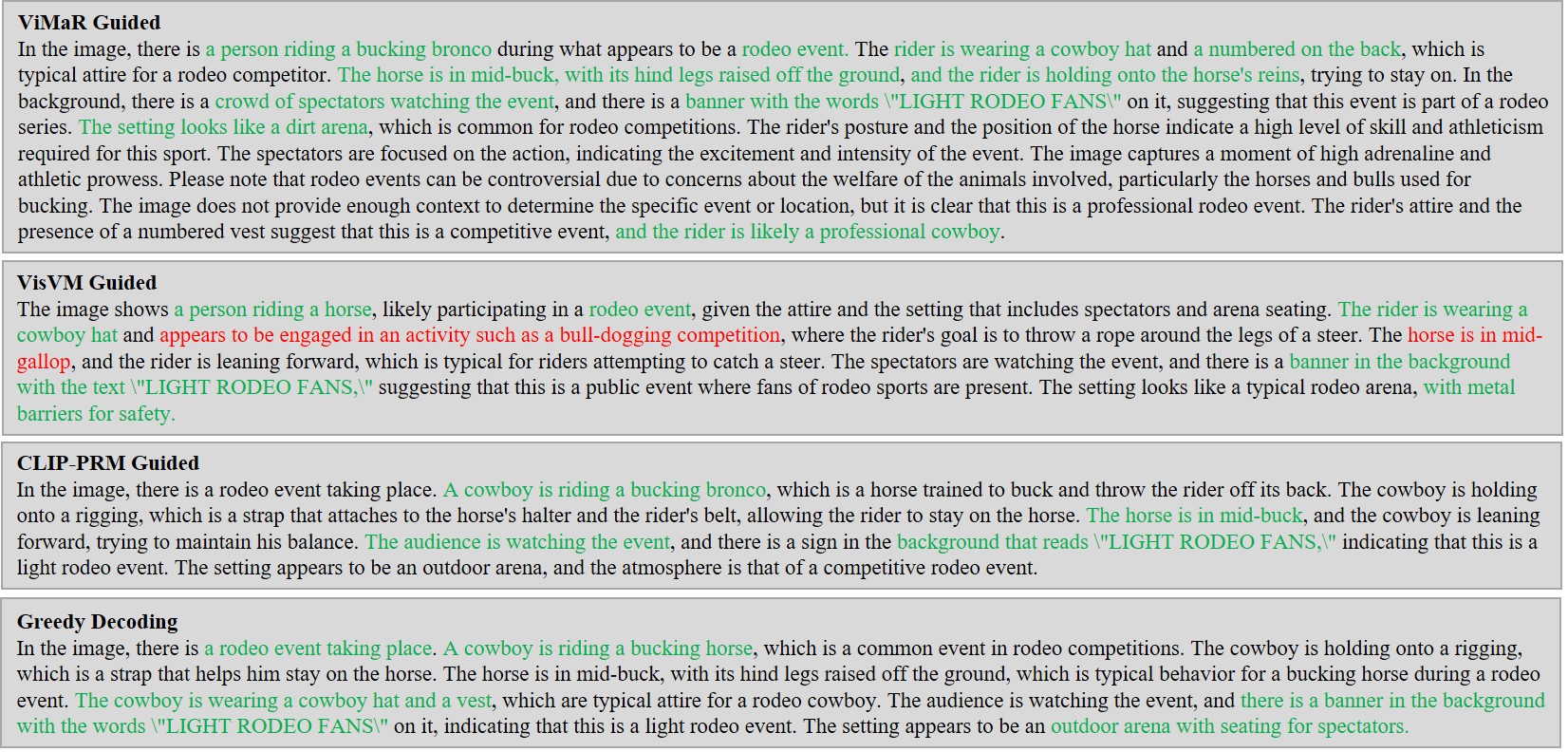}
    \caption{Case study 9: ViMaR and baseline captions.}
    \label{fig:supp9}
\end{figure}

\clearpage
\begin{figure}
    \centering
    \includegraphics[width=1.0\linewidth]{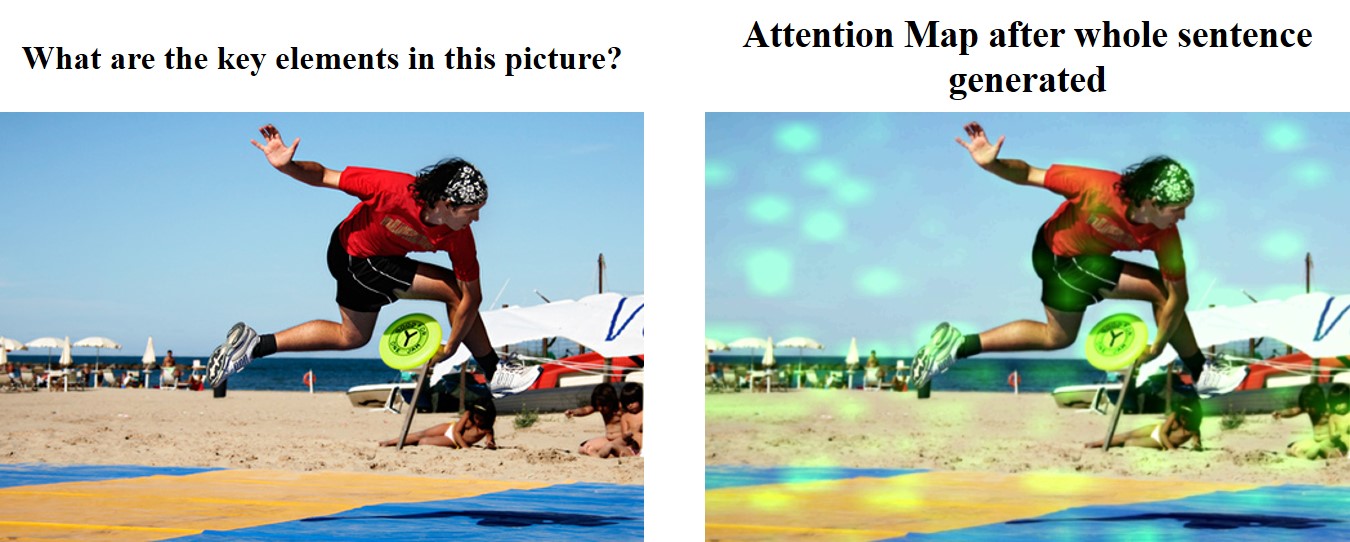}
    \hfill
    \includegraphics[width=1.0\linewidth]{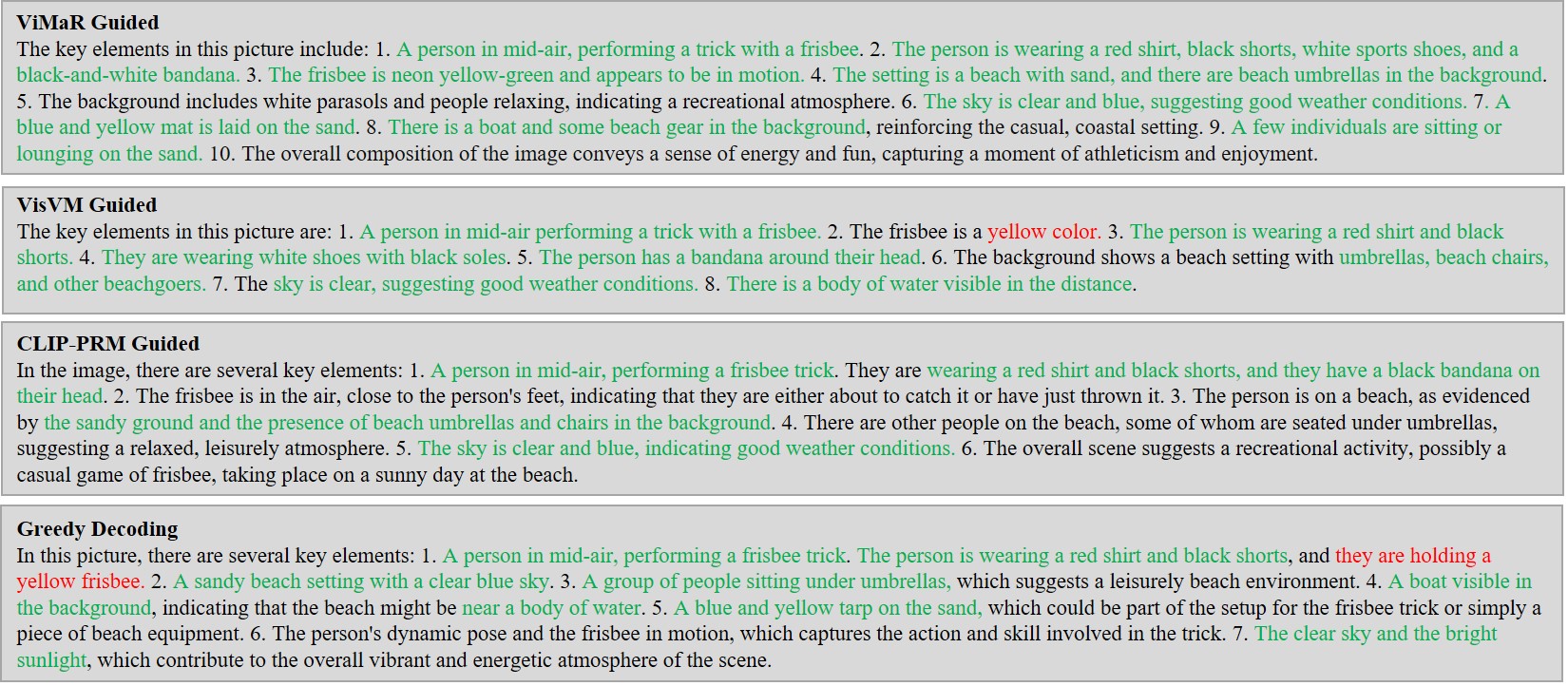}
    \caption{Case study 10: ViMaR and baseline captions.}
    \label{fig:supp10}
\end{figure}

\clearpage
\begin{figure}
    \centering
    \includegraphics[width=1.0\linewidth]{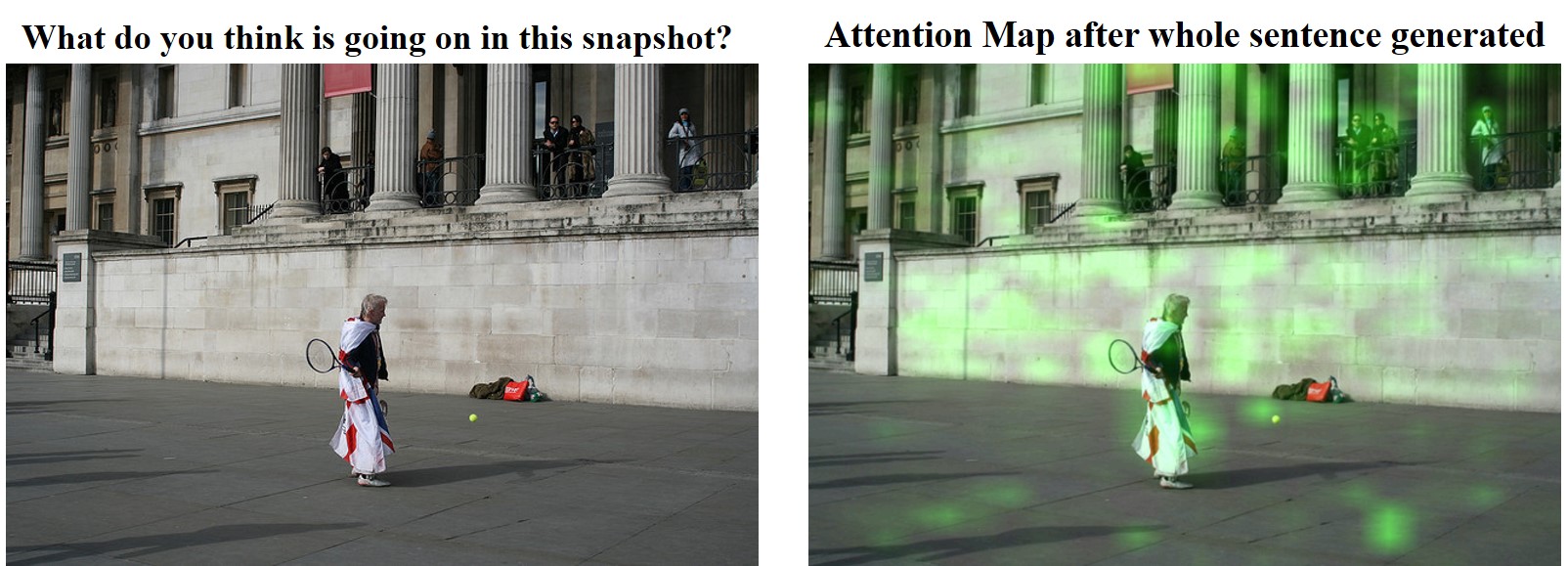}
    \hfill
    \includegraphics[width=1.0\linewidth]{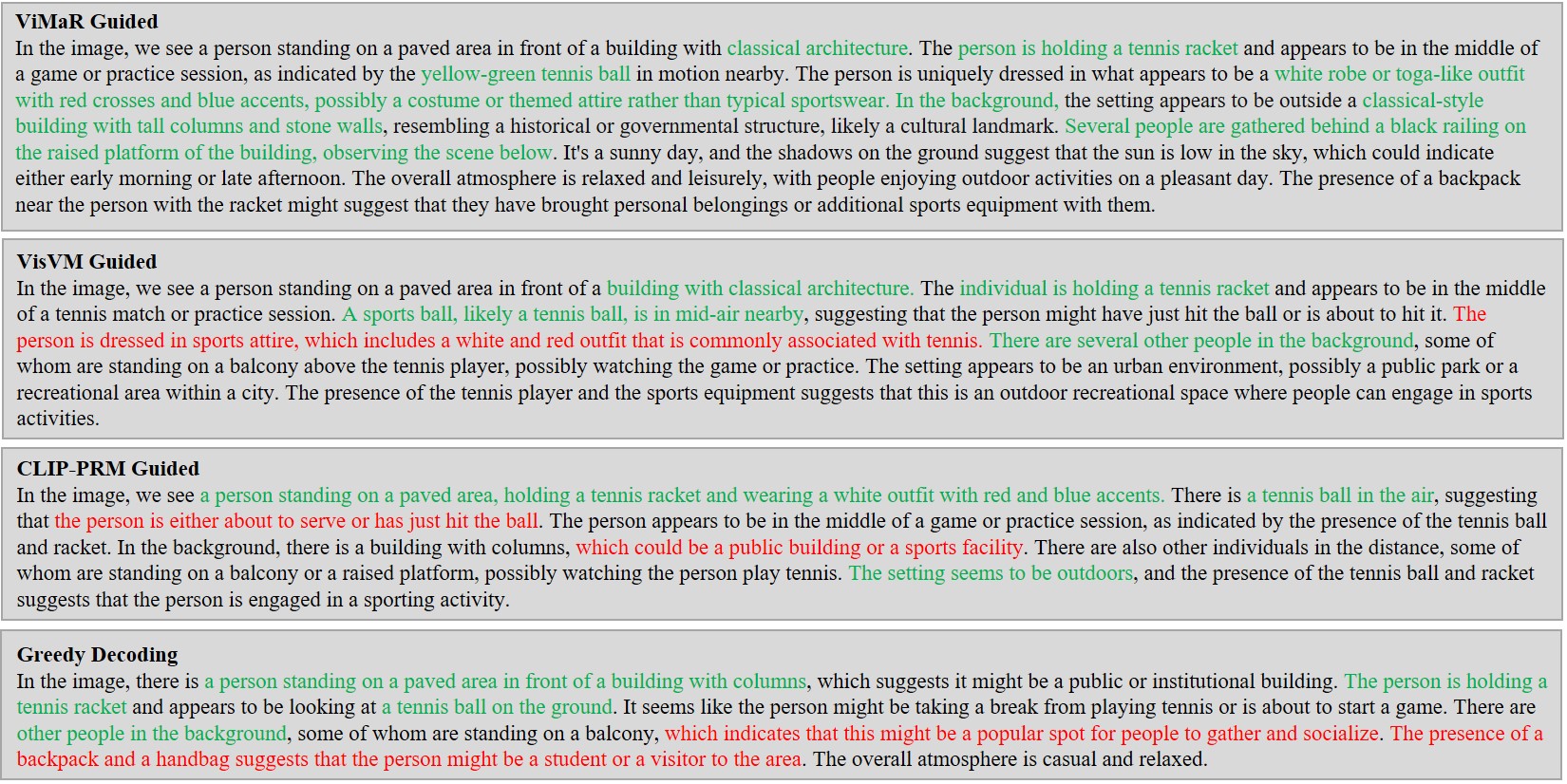}
    \caption{Case study 11: ViMaR and baseline captions.}
    \label{fig:supp11}
\end{figure}

\clearpage
\begin{figure}
    \centering
    \includegraphics[width=1.0\linewidth]{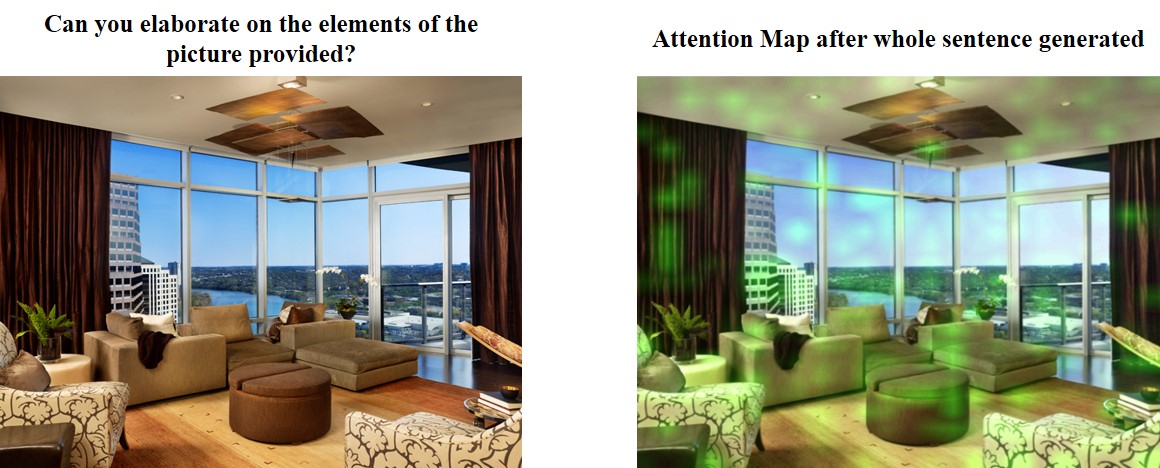}
    \hfill
    \includegraphics[width=1.0\linewidth]{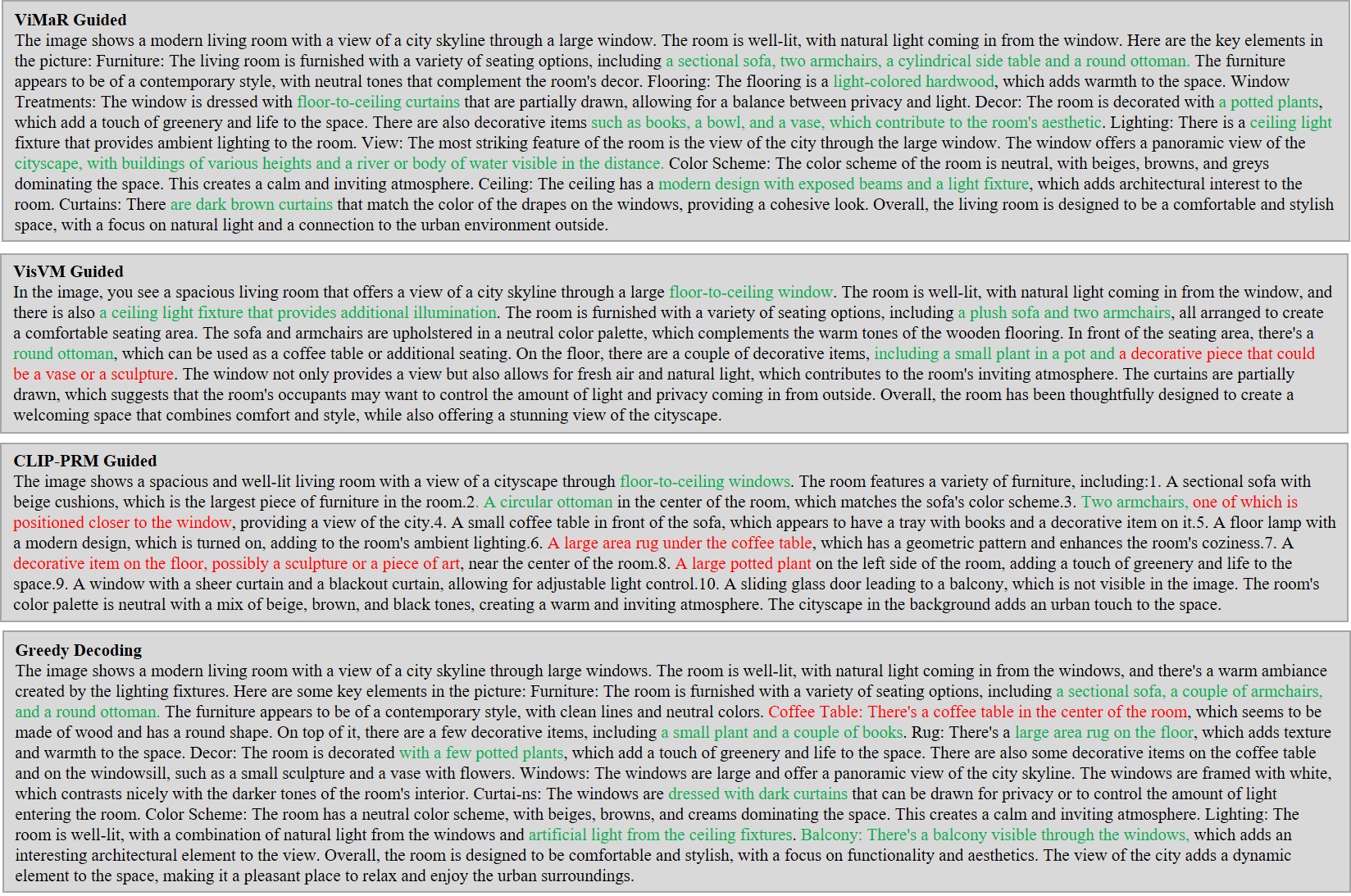}
    \caption{Case study 12: ViMaR and baseline captions.}
    \label{fig:supp12}
\end{figure}

\clearpage
\begin{figure}
    \centering
    \includegraphics[width=1.0\linewidth]{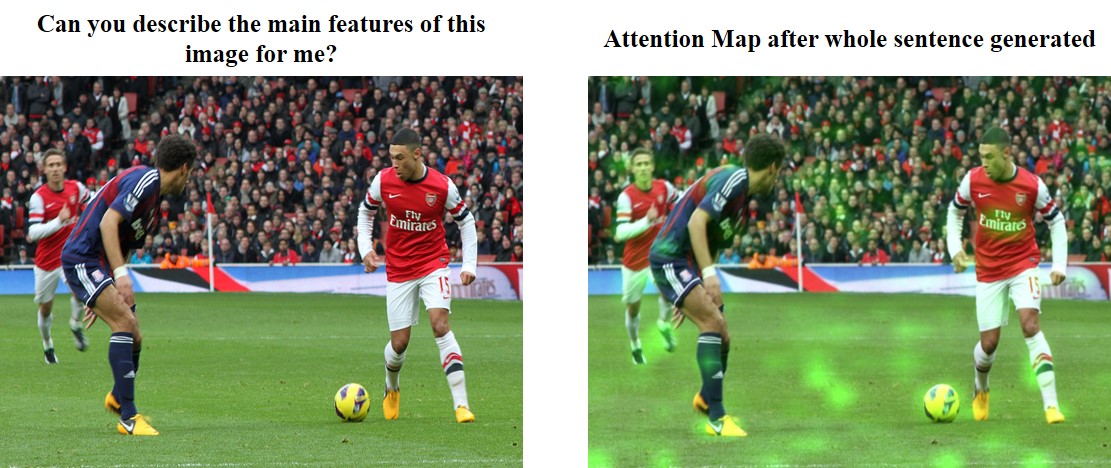}
    \hfill
    \includegraphics[width=1.0\linewidth]{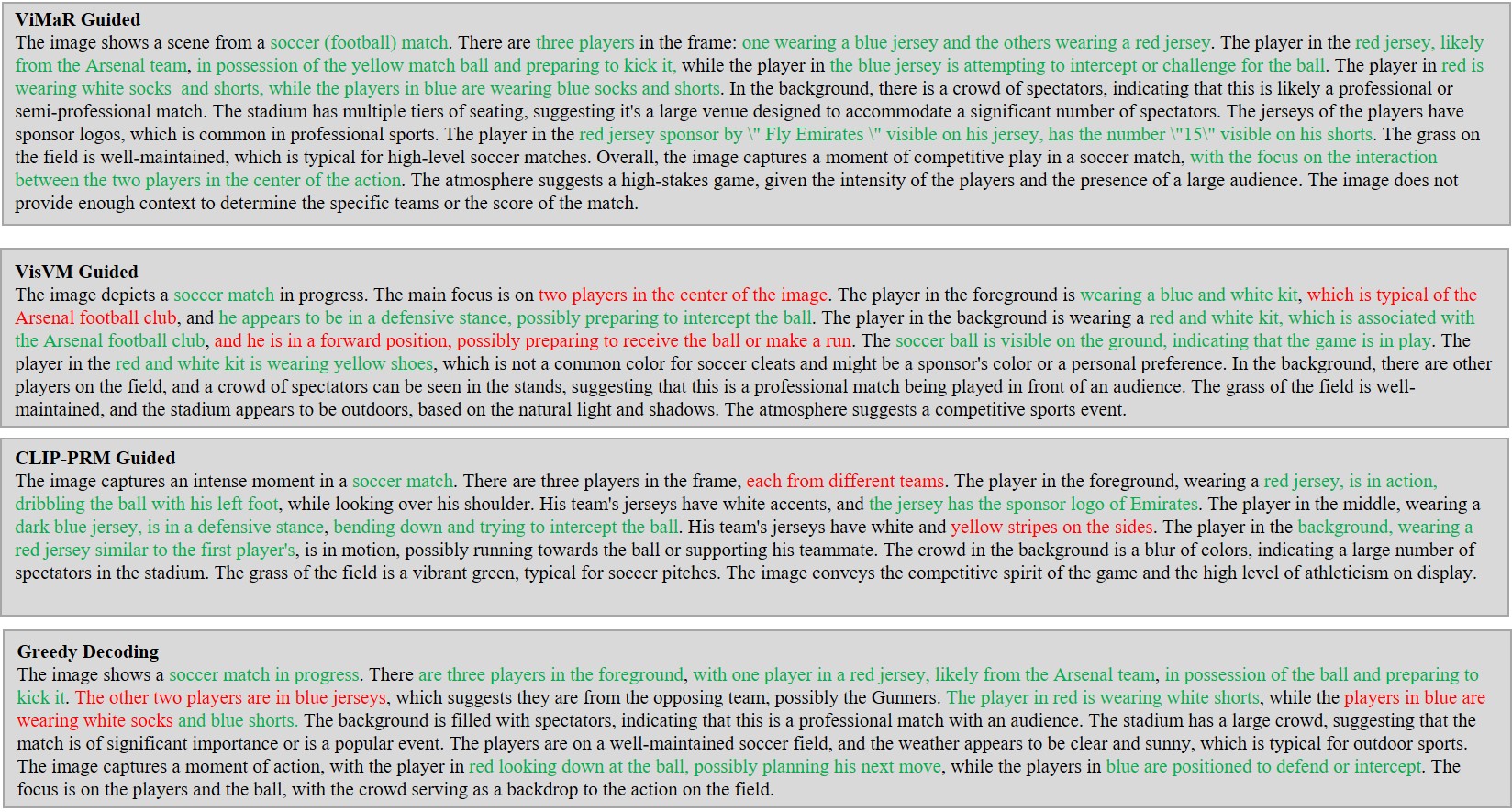}
    \caption{Case study 13: ViMaR and baseline captions.}
    \label{fig:supp13}
\end{figure}

\clearpage

\end{document}